# DeepAtlas: a tool for effective manifold learning

Serena Hughes[1,2], Timothy Hamilton[1,2], Tom Kolokotrones[3] and Eric J. Deeds[1,2,4,*]

[1]Institute for Quantitative and Computational Biosciences, University of California, Los Angeles,
[2]Bioinformatics Interdepartmental Program, University of California, Los Angeles,
[3]Department of Epidemiology, Harvard School of Public Health,
[4]Department for Integrative Biology and Physiology, University of California, Los Angeles

[*] To whom correspondence should be addressed: deeds@ucla.edu

## Abstract

Manifold learning builds on the "manifold hypothesis," which posits that data in high-dimensional datasets are drawn from lower-dimensional manifolds. Current tools generate global embeddings of data, rather than the local maps used to define manifolds mathematically. These tools also cannot assess whether the manifold hypothesis holds true for a dataset. Here, we describe DeepAtlas, an algorithm that generates lower-dimensional representations of the data's local neighborhoods, then trains deep neural networks that map between these local embeddings and the original data. Topological distortion is used to determine whether a dataset is drawn from a manifold and, if so, its dimensionality. Application to test datasets indicates that DeepAtlas can successfully learn manifold structures. Interestingly, many real datasets, including single-cell RNA-sequencing, do not conform to the manifold hypothesis. In cases where data is drawn from a manifold, DeepAtlas builds a model that can be used generatively and promises to allow the application of powerful tools from differential geometry to a variety of datasets.

## Introduction

Over the past several decades, the collection and analysis of high-dimensional data has emerged as a major theme across widely disparate fields including agriculture (1), civil engineering (2), environmental studies (3), finance and economics (4), sales (5), and many, many more. High-dimensional data is particularly an issue for single-cell genomics, such as single-cell RNA-sequencing (scRNA-seq), where we simultaneously measure the expression levels of thousands of genes across thousands to millions of single cells (6, 7). The sheer size of these datasets poses significant challenges for their analysis (8, 9).

While high-dimensional, these data are often thought of as being sampled from a much-lower dimensional manifold (10–12). This has given rise to the so-called "manifold hypothesis," which holds that high-dimensional datasets can be productively analyzed and understood in terms of the lower-dimensional manifold structures they contain (13, 14). This has given rise to "manifold learning," which attempts to develop algorithms to find these manifold structures. Current manifold learning tools are focused on dimensionality reduction, and work by generating a lower-dimensional global representation of the data (15–18). Perhaps the most well-known example of such an algorithm is Principal Component Analysis (PCA), which works by identifying a set of orthogonal components whose linear combination can capture the maximal variation within the data in question (19). Non-linear tools like t-SNE and UMAP have also become extremely popular for dimensionality reduction and visualization of data, particularly in the context of single-cell genomics studies (20–23).

As an example, consider the S-curve, which is a classic example of a 2-dimensional manifold embedded in 3-dimensional space (Fig. 1A). This dataset can easily be globally embedded into 2 dimensions, simply by uncurling that flat plane (see Methods). Application of UMAP to this dataset with the default parameters unfolds the data into 2D but introduces a set of holes or "tears" into the data. In contrast, a sphere (Fig. 1B) is also a 2-dimensional manifold, but it is mathematically impossible to embed the sphere in 2 dimensions. Application of tools like UMAP to the sphere generate an even more distorted representation, with very large tears in the dataset. These tears are problematic, in that they severely disrupt the structure of the data: points that are near neighbors of one another in the original dataset can end up on completely opposite sides of the lower-dimensional representation (Figs. 1A and 1B).

We recently developed a method to quantify this type of topological distortion and used it to show that a large set of "manifold learning" tools cannot generate effective lower-dimensional embeddings (24). We found this was true both for simple synthetic data, such as hyperspheres, and for real data from single-cell genomics (25, 26). This suggests that, while potentially useful for visualization, currently-available manifold learning tools are not able to effectively learn manifold structures from data. This is in part due to the focus that these tools have on generating *global* embeddings. As described below, the mathematical definition of a manifold is *local* in nature, but that fact is not directly leveraged by any currently-available manifold learning tool.

In a practical sense, when presented with data, we may have two questions. First, is the data actually sampled from a lower-dimensional manifold? And, if so, can we generate a mathematical model of that manifold structure? Here, we present DeepAtlas, a tool that effectively answers both of those questions. The first step in this analysis pipeline involves assessing whether or not the data is likely drawn from a manifold, since we can hardly learn a manifold structure if none is present in the data. This step also allows us to determine the dimensionality of the manifold, if one is present. The second step of the pipeline involves using a deep neural network to learn a model of the manifold itself. We do this by learning a set of "local charts" for the data, and the collection of these charts forms an atlas for the manifold.

Application of the first part of the pipeline to synthetic data demonstrates that DeepAtlas accurately estimates the dimensionality of manifolds, even when the dimensionality of the data is high relative to the number of sampled data points. Interestingly, we find that several real datasets actually show no evidence of being drawn from a manifold, including those classically used for testing manifold learning tools and scRNA-seq data. Other real-world datasets, however, are consistent with being drawn from a manifold, including the classic MNIST test dataset of images of hand-written digits. For both synthetic and real data, we demonstrate that DeepAtlas can effectively learn a differentiable model from this data. We show that this model can be used generatively to produce new data points from the original manifold by sampling from the local representations of the data. To our knowledge, this work represents the first attempt to actually directly apply the mathematical definition of a manifold to study data.

**Results**

Overview of DeepAtlas

As mentioned in the introduction, high dimensional data is often analyzed based on the "manifold hypothesis," which posits that high-dimensional data is drawn from a lower-dimensional manifold (see the examples of the S-curve and sphere in Figs. 1A and 1B). In discussing the manifold hypothesis further, it is helpful to revisit the mathematical definition of a manifold. Our discussion here is not meant to be completely formal, but is consistent with more rigorous treatments (11, 12, 27, 28). We say a set $M$ is an $n$-dimensional manifold if and only if there exists a covering of open subsets of $M$ such that, for each open subset $U_i$ in this covering, there exists a homeomorphism between $U_i$ and an open subset of Euclidean space $V_i \subset \mathbb{R}^n$. Recall that a set $U_i$ is homeomorphic to a set $V_i$ when there is a continuous, invertible function between them such that the inverse is also continuous; we call this function $\phi_i$ for each member of this covering. Intuitively, this means that there is a way to "locally" represent regions of the manifold as an $n$-dimensional Euclidean space (Fig. 1C). A great example of this is the surface of the Earth. Globally, the earth is obviously spherical, but it is possible to make a completely flat map of a city like Los Angeles. Every point in Los Angeles corresponds to a point on the map (e.g., think of finding a specific address on a map); that correspondence represents the action of the function $\phi_i$. As a result of this obvious correspondence with maps of locations on the earth, these $\phi_i$ functions are often referred to as "charts" or "coordinate charts". The collection of the set of charts $\{\phi_i\}$ and their corresponding open subsets $\{U_i\}$ of $M$ forms an "atlas," similar to how a collection of local maps can form an atlas of the earth.

It is often helpful to consider a "differentiable manifold," which imposes an additional requirement. Specifically, consider two sets $U_i$ and $U_j$ in the covering of open sets in $M$, and say they have a non-trivial intersection ($U_i \cap U_j \neq \emptyset$). If the manifold is differentiable, then for any point in the image $\phi_i(U_i \cap U_j)$ of this intersection, $\phi_j \circ \phi_i^{-1}$ must be differentiable; here, we will take these functions to be smooth, meaning infinitely differentiable. These overlapping regions are often referred to as "transition regions." In the example of using a collection of local 2D maps to describe the 3D globe of the Earth, these transition regions are the areas shown at the edges of multiple maps so that you can cross from one local region to the next while navigating.

The collection of algorithms referred to as "manifold learning" universally seek to obtain global embeddings of the data (11); for instance, UMAP attempts to globally "unfold" datasets like the S-curve globally into a 2-dimensional representation (Fig. 1A). However, many of these algorithms actually fail to generate reasonable embeddings. For instance, the popular manifold learning algorithm UMAP performs fairly well on the S-curve, but introduces tremendous distortion when applied to a non-trivial manifold like the sphere (Fig. 1B). Indeed, we found that manifold learning algorithms generally fail to generate effective embeddings, even when presented with relatively trivial problems like embedding a hypersphere (24). This suggests a need to develop algorithms that leverage the definition of a manifold as a local representation.

Here we propose a new algorithm, DeepAtlas, as a solution to this problem. The approach has three main steps: 1) generate local neighborhoods, 2) determine the embedding dimension and generate local embeddings, and 3) learn the functions $\{\phi_i\}$ to construct an atlas between the high and low dimensional representations of each local neighborhood (Fig. 1C).

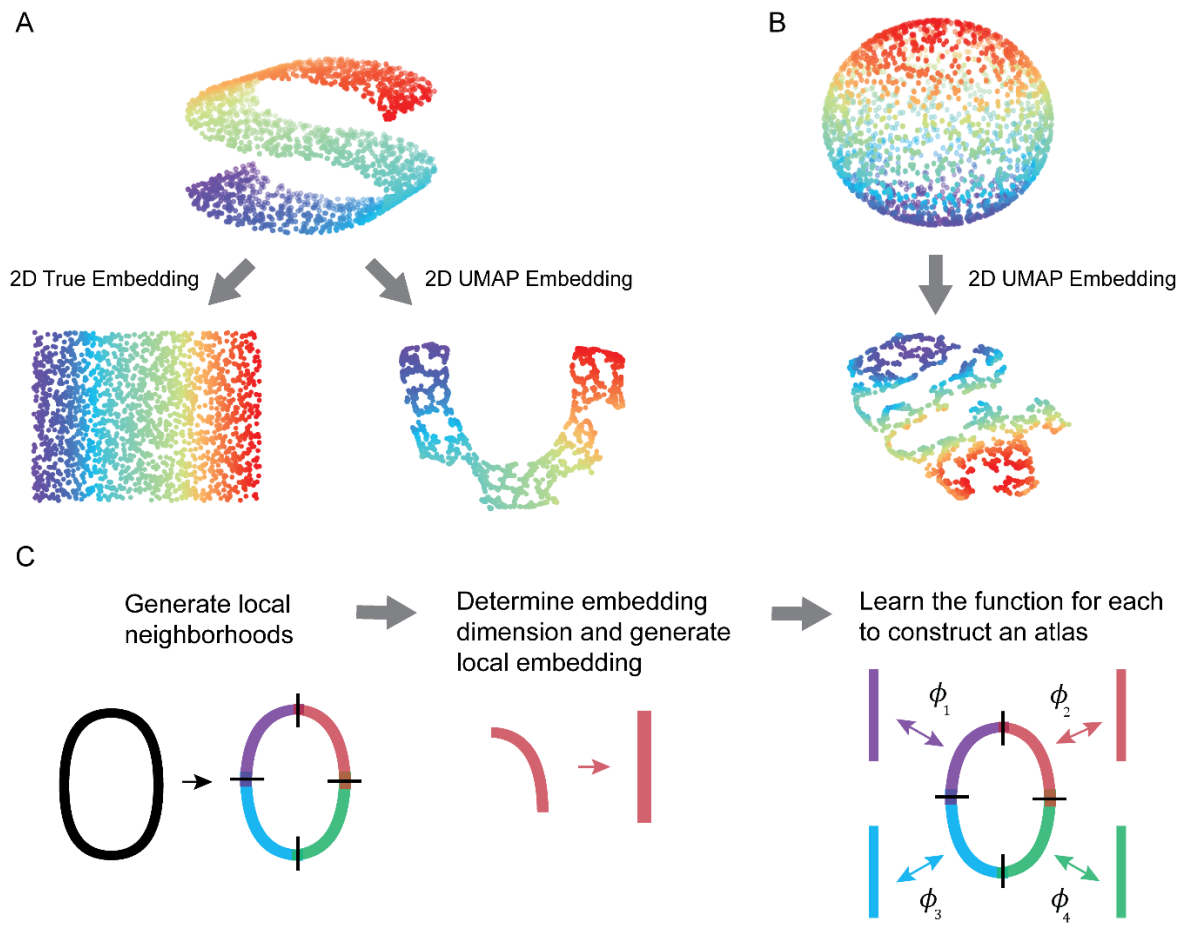

**Fig. 1. Learning manifolds.** (**A**) The 3D S-curve dataset and true 2D embedding (left) and UMAP embedding (right). (**B**) A spherical dataset and its UMAP embedding. (**C**) Schematic of the steps of DeepAtlas applied to a 2D oval shape.

This process allows DeepAtlas to approach the data locally to minimize distortion, identify the dimension of a manifold if one exists, and learn the functions that map between local embeddings in that lower dimension and the original dataset.

Clustering

The first step in DeepAtlas is to determine the covering of subsets, which we do by generating "local neighborhoods" in the original data (Fig. 1C). Here we use the $k$-means clustering algorithm, which groups points based on the average distance to the cluster's center. We chose to focus on $k$-means because it generates local neighborhoods that correspond to our geometric intuition from common examples like the sphere or S-curve, and because we found it works well on the example cases described below. Importantly, DeepAtlas is designed to be modular and any other clustering algorithm can be substituted as needed.

Once the data has been partitioned into $k$ clusters, we determine the transition regions between the clusters. To do this, we first compute the $l$ nearest neighbors of all the points in the dataset. A transition point is defined as a point that has at least one of these $l$ nearest neighbors in a different cluster. We then expand the original clusters as follows: for every transition point, we add each of its neighbors from other clusters to the original cluster. This means that our clusters have overlapping sets of points on their borders. As an example, we used $k$-means clustering to partition the S-curve into $k = 5$ clusters (Fig. 2A). After adding the transition points to the relevant clusters with $l = 10$ nearest neighbors (which are colored black in Fig. 2A), it is clear how these local regions can be "stitched together" by the transition regions between them. Note that $k$ and $l$ are parameters that can be set by the user; some discussion of how to determine meaningful values of these parameters, particularly $k$, can be found below.

Generating local embeddings

The next step in the pipeline involves generating the lower-dimensional embeddings of these local regions. To do this, it is helpful to introduce our previous work, which focused on understanding to what extent an embedding introduces *topological distortion* into the data (24). To quantify this distortion, we developed a statistic based on the Jaccard distance, which measures dissimilarity between the $h$ nearest neighbors of a point in the original $m$-dimensional dataset and the $h$ nearest neighbors of that same point when it is compressed into $n$ dimensions, where $m > n$. Let $A_i$ be the set of $h$ nearest neighbors of a given point $i$ in the original data and $B_i$ be the set of $h$ nearest neighbors of $i$ in the lower-dimensional representation (Fig. 2B). The Jaccard distance for point $i$ is defined as:

$$J_i = \frac{|A_i \cup B_i| - |A_i \cap B_i|}{|A_i \cup B_i|} \quad (1)$$

where $|X|$ represents the cardinality of the set $X$. The larger the Jaccard distance, the more dissimilar the neighborhoods are: when $J_i = 0$ the neighborhoods are *identical* and there is no distortion, and when $J_i = 1$ the neighborhoods of the point are completely different. Formally, a lower-dimensional representation that is actually an embedding should have no distortion, so we should have $J_i = 0$ at each point. A simple way to get a sense for the overall distortion in a dataset is to average the Jaccard distance across all the points in the dataset. This produces a statistic we call the "Average Jaccard Distance" or AJD:

$$AJD = \frac{1}{N} \sum_i J_i \quad (2)$$

where $N$ is the total number of points in the dataset (Fig. 2B).

We tested a wide variety of dimensionality reduction algorithms on simple manifolds like the hypersphere and found that nearly every algorithm we tested introduced high levels of distortion (24). For instance, application of UMAP to the S-curve in Figs. 1A and 2B results in an AJD of 0.35, indicating a high level of distortion even for this simple case. Indeed, the only approach that we found that could *reliably* reduce the dimensionality of data with close-to-0 distortion in the known minimal embedding dimension for hyperspheres was PCA (19, 24), which was effective regardless of the underlying dimensionality of the hypersphere from which the data was drawn. As such, in this work, we focus on using PCA to generate our local embeddings. As mentioned above, DeepAtlas is entirely modular, and can be used with any linear or non-linear dimensionality reduction tool.

We first tested this approach on the S-curve and sphere, which we know both have a local dimension of 2. We found that PCA generated very low-distortion embeddings for the *local* clusters in these data, even though the global structures could not be embedded in 2D without very high levels of distortion (Fig. 2C). We next considered the "Swiss Roll," another locally 2D manifold frequently used for testing manifold learning algorithms (Fig. S1, 29). Interestingly, we found that setting $k=5$ resulted in high levels of distortion (Fig. 2C); visual inspection of the 2D embeddings showed that the problem in this case was caused by the fact that, at $k=5$, the clusters were not confined to points on the manifold, but rather included points that were close in 3-dimensional space but not neighbors on the surface of the Swiss Roll (Fig. S1A and S1C). We progressively increased the value of $k$ until the local embeddings had low AJD (Fig. S1B and S1D). This demonstrates that the strategy of using PCA for local embeddings works also with the Swiss Roll, as well as highlighting how the AJD can be used to guide the choice of critical parameters.

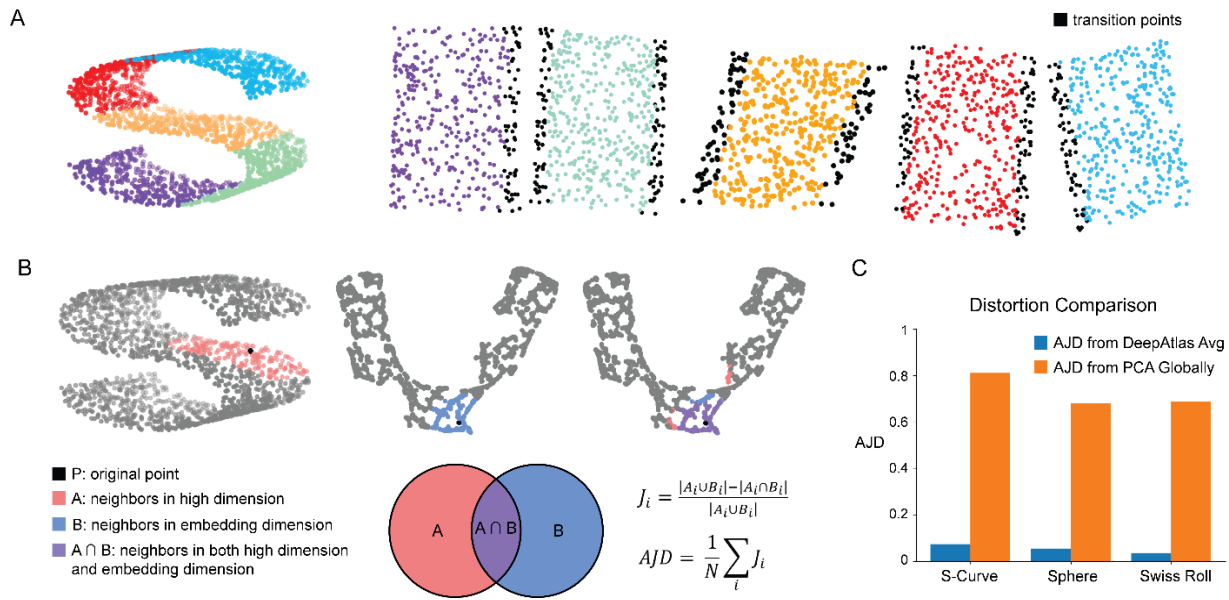

**Fig. 2. DeepAtlas Leverages Local Neighborhoods.** (**A**) 3D S-curve example partitioned into $k=5$ clusters. Each cluster has transition points included in both neighboring clusters, highlighted in black. Clusters are reduced to 2D with PCA and aligned to show how they can be stitched together. (**B**) Visual explanation and mathematical definition of the Average Jaccard Distance (AJD), a metric used to quantify topological distortion introduced by dimensionality reduction. (**C**) Comparison of the AJDs of global PCA and average AJDs across local neighborhoods generated by DeepAtlas.

In general, however, we will not know *a priori* what the local dimension $n$ is for any given dataset we encounter. The approach we take thus involves applying PCA to each of the $k$-means clusters and varying the number of principal components (PCs) used to represent the data from 1 to $m$, the original dimensionality of the data. The resulting plot of the AJD vs. local embedding dimension provides a straightforward way to determine the value of $n$. To test this approach in higher dimensions, we considered the 9-dimensional hypersphere ($S^9$), which can be *globally embedded* in 10 dimensions or more ($\mathbb{R}^{10}$ or higher). We used a simple computational approach we developed in our previous work to sample 5,000 data points with uniform probability from the surface of the 9-dimensional hypersphere; note that each point sampled in this way is just a vector in $\mathbb{R}^{10}$ (see Methods for further details, 24). We can trivially embed this manifold in $\mathbb{R}^{20}$ by appending 10 zero elements to each vector. While this is obviously an extremely simple example, we note that all available "manifold learning" tools we tested, including popular tools like t-SNE and UMAP, were unable to identify a low-distortion embedding for this and similar examples, even when attempting to embed from $\mathbb{R}^{20}$ into $\mathbb{R}^{10}$. Indeed, PCA was the only tool we found that could achieve success in this case (24).

To test our approach for identifying the correct *local* embedding dimension, we first applied $k$-means clustering to the data, using $k = 10$ as a starting point. This resulted in an approximately uniform set of cluster sizes (Fig. S2). We then calculated the AJD for the PCA embedding as a function of the embedding dimension for the individual clusters. As can be seen from Fig. 3A, there is a very distinct behavior here—the clusters' AJD is close to 0 at a local dimension of 9, and the individual clusters generate almost exactly the same curve. We applied this approach to several other example hyperspheres, ranging in dimensionality from 5 to 1000, and found consistent performance—PCA was always able to identify the correct local embedding dimension (Fig. 3B, Fig. S3).

We then applied this approach to more realistic datasets that are often used as examples for testing unsupervised and manifold learning algorithms (30). This includes a dataset of biomarkers from the Pima Native American peoples, features of houses related to housing prices, and physical characteristics of wine and wheat seed samples (Fig. 3B and 3C), as well as several others (Fig. S4). In each case, we chose an initial value of $k$ such that the smallest clusters would never have fewer than 50 points. Since these datasets vary significantly in size (from $N = 210$ for the Wheat dataset to $N = 4,898$ for the Wine dataset), the value of $k$ for our initial tests also varied (Figs. 3C–F). Interestingly, we saw two distinct classes of behaviors for these datasets. In the Pima and Housing datasets, we see that individual clusters vary wildly in terms of AJD vs. embedding dimension using PCA. For instance, in the 8-dimensional Pima dataset, some clusters achieve relatively low distortion (less than 0.2) at around 6 dimensions, while others require all 8 dimensions to achieve similarly low distortion (Fig. 3C). The case for the Housing dataset is similar (Fig. 3D). In the Wine and Wheat datasets, however, we see much more uniform behavior across clusters; the 12-dimensional Wine dataset is (locally) 11-dimensional (Fig. 3E), while the Wheat dataset is approximately 4-dimensional (Fig 3F).

These findings indicate that the local neighborhoods of the Pima and Housing datasets do not have a consistent dimensionality. Formally, a dataset might be drawn from a union of manifolds, that is, disjoint "pieces" of different dimensionality: consider, for instance, a dataset consisting of the union of a circle and a sphere (Fig. S5). Each connected component of this dataset is its own manifold with a single local dimensionality; if the local dimensions are the same, we can technically consider the entire dataset as a manifold. This suggests that, in addition to exploring the dimensionality of these local neighborhoods, it is also important to consider whether the dataset itself might consist of the union of different manifolds.

DeepAtlas offers two complementary approaches to determining the underlying connectivity of the dataset. The first is generated by the "transition points" between the clusters we obtain. We can use transition points to represent any dataset as a graph, where the nodes are the clusters and we place an edge between two clusters if they have transition points shared between them (in other words, if one of the nearest neighbors of a point from one cluster is in the other). A picture of this graph is provided for each of the datasets in Fig. 3; note that here, the weight or "width" of each edge is proportional to the number of transition points between the two clusters. This approach is similar to the PAGA algorithm developed by Wolf, Theis, and co-workers for visualization and analysis of scRNA-seq data (31). DeepAtlas also provides a complementary method for finding distinct connected components in the data, which is based on a graph-based approach that we call $\epsilon$ networks (also known as random geometric graphs, Fig. S6, 32).

The hypersphere datasets we generated clearly form a single connected component by construction, and that is reflected in the graph of connections between local neighborhoods for these datasets (Figs. 3A and 3B). The Pima dataset forms a single connected component regardless of whether we use the transition point graph (Fig. 3C) or the $\epsilon$ network approach (Fig. S6C). As mentioned above, however, the local neighborhoods of a connected component of a manifold must have the same dimensionality, and so this stage of DeepAtlas pipeline strongly suggests that the Pima data cannot be modeled as a manifold. Interestingly, we find that the Housing dataset almost certainly comprises two separate components (Fig. 3D and Fig. S6D); in this case, however, neither of those components have a consistent local dimension (Fig. 3D). So, as with the Pima data, the Housing data does not represent a manifold structure, at least as revealed in the current DeepAtlas analysis.

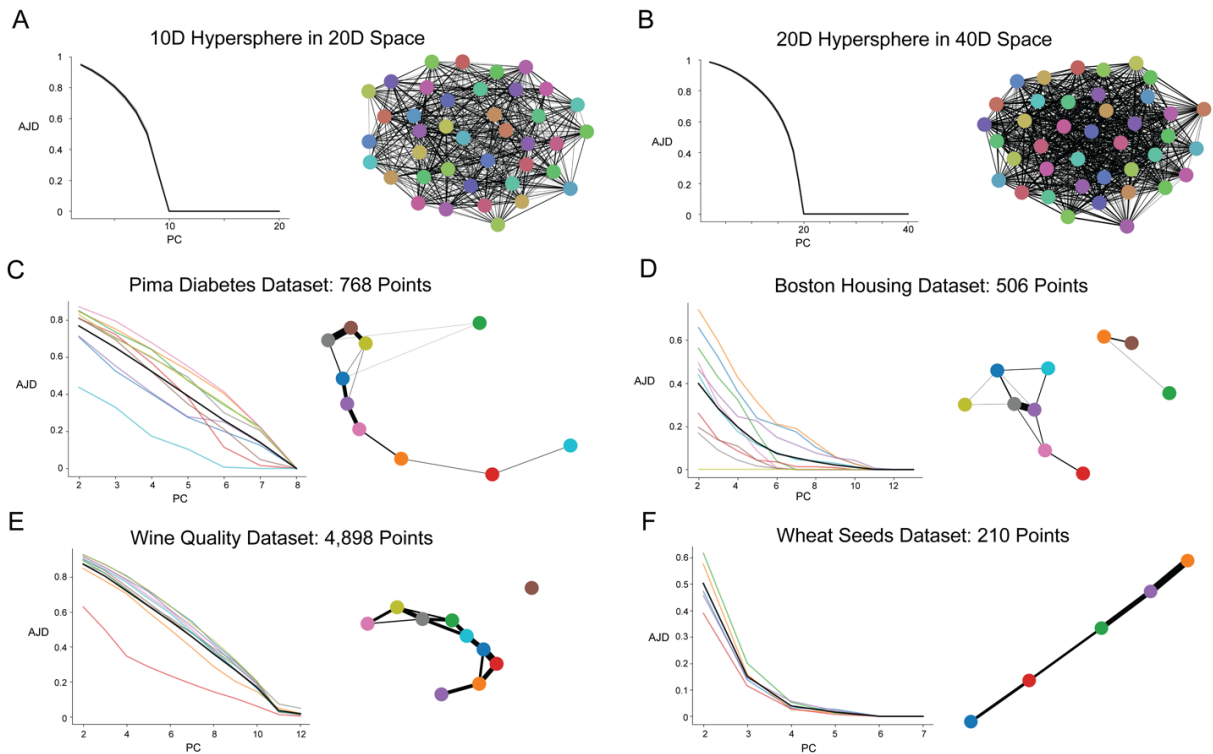

**Fig. 3. DeepAtlas Performs Well on Test Datasets.** AJD vs. PCA embedding dimension plots, where each gray or colored line is a cluster and the black line is the average. (**A**) 10D hypersphere in 20D space. (**B**) 20D hypersphere in 40D space. (**C**) The Pima diabetes dataset. (**D**) The Boston housing dataset. (**E**) The wine quality dataset. (**F**) The wheat seeds dataset. Note that we do not see indication of a consistent lower-dimensional manifold in C or D.

Both the Wine and the Wheat datasets, however, do seem to have a consistent manifold structure (Figs. 3E and 3F). In the case of the Wine dataset, there is one local cluster that has no connection with the rest of the data. Regardless, the connected component of the data has a consistent dimension of 11 (Fig. 3E). Interestingly, the Wheat dataset forms a simple line-like structure where each local neighborhood is approximately 4-dimensional (Fig. 3F). Using the approach shown in Fig. 3, users can simultaneously evaluate whether their data is likely taken from a manifold, and, if so, determine the local dimensionality of that manifold.

Learning the Atlas

The final step of DeepAtlas is to learn the functions $\{\phi_i\}$ to construct an atlas. We first note that the individual functions that comprise a smooth atlas are often taken to be *diffeomorphisms*, meaning that they are invertible, continuous, and smooth in both directions. For the "forward" direction, i.e. mapping between the high- and low-dimensional spaces, we just use PCA itself.

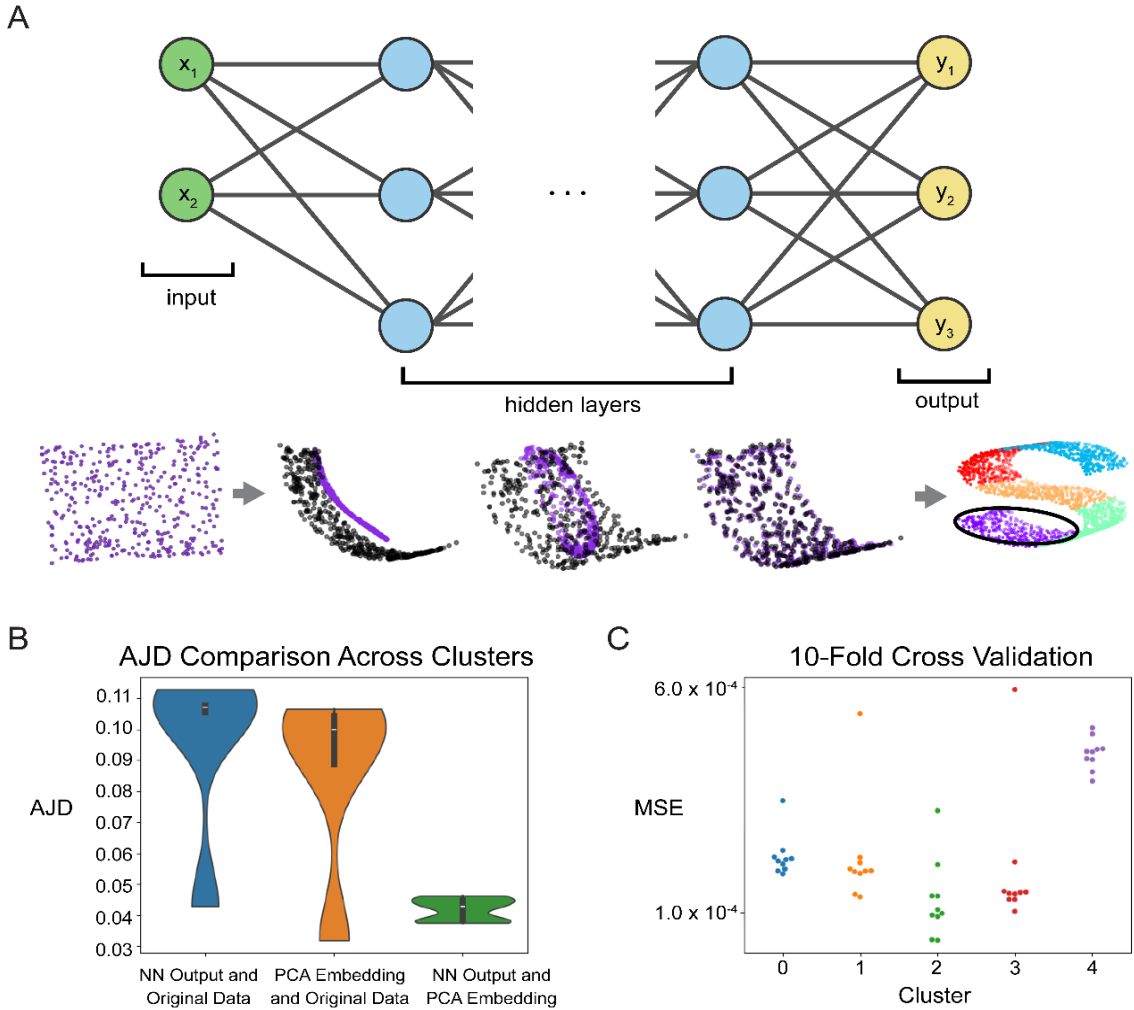

**Fig. 4. Learning Inverse Charts.** (**A**) Schematic of the neural network structure used by DeepAtlas to approximate the inverse of each chart. Beneath we show an example of training the neural network to map back to the original high dimensional data. (**B**) AJD comparison to evaluate the performance of the neural network model on the S-curve data. The distributions in the violin plot are taken across the 5 $k$-means clusters. (**C**) 10-Fold Cross Validation results to further evaluate the neural network performance, measured using Mean Squared Error (MSE).

PCA is lossy, and so does not possess an exact inverse, so we use a deep neural network to learn the inverse $\phi_i^{-1}$ for each chart (Fig. 4A). This function takes in the position of a point in the PCA embedding and attempts to predict the original position of that point in the starting dataset. To train this network, we use the standard Mean Squared Error (MSE) approach, where we quantify the error based on the average square distance between the output values of the network and the position of that point in the original high-dimensional space. Given this architecture, we can use standard tools for training neural networks, for instance Stochastic Gradient Descent (SGD) methods implemented in Python, to train the networks themselves (33).

      Of course, the models' usefulness hinges on the accuracy of their output embeddings. We tested multiple versions of this particular architecture with a varying number of layers on a variety of test datasets, all of which yielded similarly good results. For instance, applying our prototype algorithm to the S-curve data, the model achieved a low Mean Squared Error (MSE) with 10 hidden layers and 10,000 epochs of SGD (Fig. S7). To evaluate performance, we first considered the AJD between the original data (i.e. the points on the S-curve manifold, Fig. 4B)

and the various low-dimensional representations. In particular, we see low AJD values between the neural network output and the original data, indicating that the training was very successful in this case. The model also consistently yields low MSE when tested with 10-fold cross-validation (Fig. 4C). We also applied trained similar models for the hypersphere data and the machine learning test datasets (namely the Wheat and Wine datasets, where we see evidence of a consistent manifold, Fig. 3E and 3F), obtaining similarly good performance (Fig. S8). We note that users can alter the network architecture and training protocol to suit any particular dataset. The combination of PCA and these learned inverses gives us an approximation of the atlas for the dataset.

scRNA-Seq Datasets Lack Manifold Structure
A major application area for manifold learning tools is scRNA-seq data. One of the main challenges of analyzing this data is its high dimensionality; raw datasets often report the gene expression of tens of thousands of genes in tens of thousands to even millions of cells. In these data, cells can be thought of as being points in a >20,000-dimensional vector space, and as a result manifold learning tools are essentially ubiquitous in this field. This includes the almost universal reliance on PCA upstream of analyses like cell type clustering, as well as a reliance on visual inspection of 2D visualizations of the data (particularly UMAP plots) to characterize and evaluate cell type clusters and patterns of gene expression (34–37). scRNA-seq data is widely thought to have a lower-dimensional manifold structure, and many pipelines (implicitly or explicitly) make an appeal to the manifold hypothesis to argue that downstream analyses can meaningfully be performed using lower-dimensional representations (36–38). Our previous work demonstrated that linear tools like PCA and non-linear tools like t-SNE and UMAP introduce tremendous topological distortion into scRNA-seq data, with AJD values typically above 0.9 (24). This and other studies (39–42) all suggest that current manifold learning techniques struggle to successfully embed or otherwise represent the actual structure of scRNA-seq data.

     We applied DeepAtlas' pipeline to explore whether scRNA-seq datasets likely conform to the manifold hypothesis. Very few (if any) analyses of scRNA-seq data are carried out on the raw count data itself. Rather, the data is subjected to a more-or-less standard pipeline of transformations before dimensionality reduction is performed; this includes normalizing the data by Counts Per Million (CPM), identifying and focusing on a subset of "Highly Variable Genes" (HVGs), and log transformation (6, 7, 36). We expect that the relevant lower-dimensional manifold should be evident from the raw data, but by applying DeepAtlas at each stage in this pipeline we were able to test how these transformations might influence the underlying structure of the data.

     There are a large number of published scRNA-seq studies to which DeepAtlas could be applied. We first chose to focus on an important dataset in which immune cells were first separated into separate cell-type groups using Fluorescence-activated cell sorting (FACS) before being sequenced (43). This allows us to apply our pipeline to groups of cells that all come from the same cell type, based on well-established celltype markers. The data shown in Figure 5 is scRNA-seq data for the Natural Killer (NK) cells. We also applied DeepAtlas to other scRNA-seq data including the full PBMC data from the same study that provided the NK data, data from cell lines that also has cells separated into natural groups (44), and simulated scRNA-seq data (Fig. S9, 45).

     The first step of DeepAtlas is to generate local neighborhoods; here we used $k$-means clustering with $k = 40$. For 40 $k$-means clusters in the raw NK cell data, we found that the

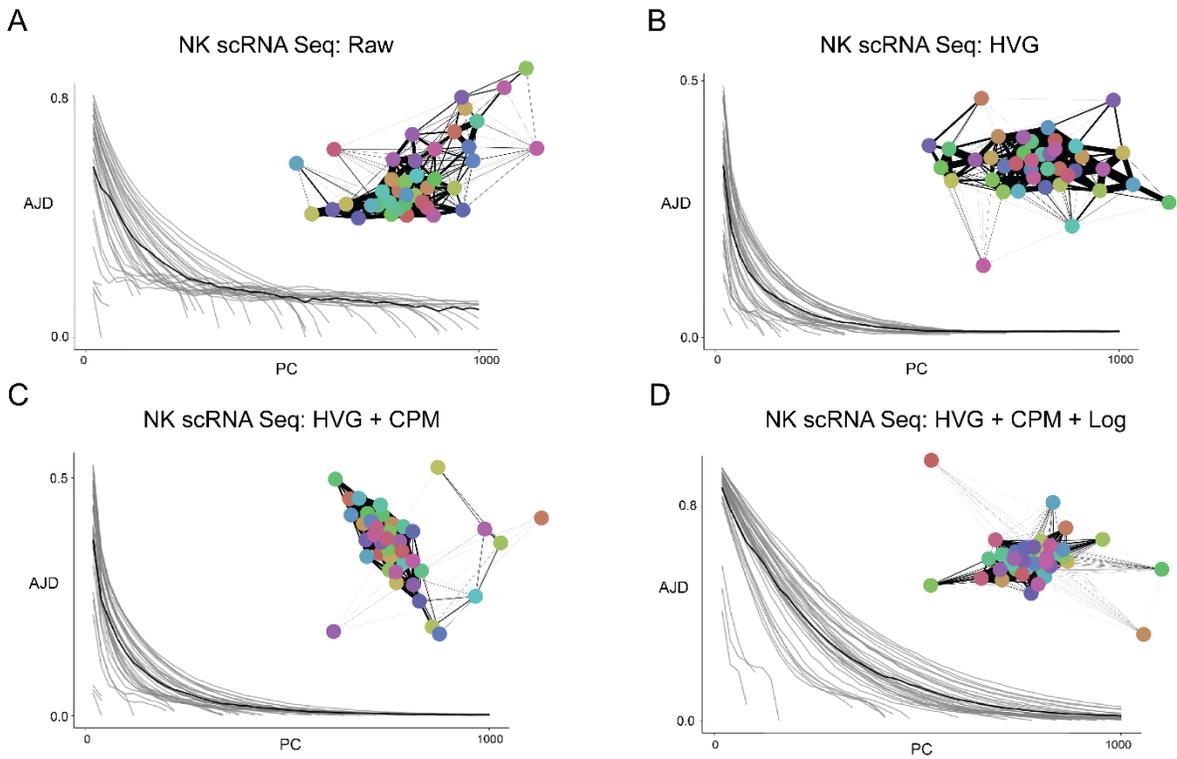

**Fig. 5. We Do Not Observe Manifolds in scRNA-seq.** PC vs AJD plots for the NK cell scRNA-seq data. All cases used $k = 40$ clusters. (**A**) Raw data. (**B**) Highly Variable Genes (HVGs). (**C**) Counts Per Million (CPM) Normalized HVGs. (**D**) Fully transformed data: CPM and log transformation applied to the HVGs. Note the lack of consistent manifolds for any of these cases, particularly D, which is the space most often used as the basis for scRNA-seq analysis.

clusters range in size from having as few as 5 cells in a cluster to as many as 1679 (Fig. S10A). We tested varying values of $k$, but the results were similar (Fig. S11, S12). The trend of widely varying cluster sizes was consistent across all stages of data transformation/preprocessing and across all datasets we considered (Fig. S10). This difference between sizes of clusters was also consistent across various values of the $k$ parameter used for $k$-means clustering (Fig. S11). These clustering results are very different compared to the other datasets we considered, where cluster size was very consistent across the various clusters (Figs. S2, S13), and may be a result of the highly uneven distribution of cells in these various transformed gene expression spaces (46).

From a practical perspective, the size of a cluster caps the number of dimensions that the smaller clusters can be reduced to; if we have only 5 cells in a cluster, then at the 5$^{th}$ PC there is essentially one component for each cell, and the dimensionality of that cluster cannot be reduced further. This results in sudden, dramatic decreases in the AJD when the PC number gets close to the number of cells in the cluster (Fig. 5). We found that the curves for each cluster do not align closely with each other; some clusters require hundreds of more PCs to reach a low AJD compared to others (Fig. 5). This indicates that, at least in space of the raw counts, the cells are not sampled from a consistent manifold (Fig. 5A). Interestingly, we found lower AJD values overall for the data when just considering the top 2000 HVGs (6, 7, 36), but the individual clusters still do not align to show a consistent local dimension (Fig. 5B). Neither CPM normalization of the data nor log transformation resulted in consistent dimensionality across clusters (Fig. 5C); indeed, after the final step of log transformation, the AJD values increase and the cluster curves visually separate even more (Fig. 5D). This trend was also seen when

DeepAtlas was run across other scRNA-seq experiments (44) and even simulated scRNA-seq data generated using the scDesign package (45) (Fig. S7). Note that, in all these cases, analysis of transition points and $\epsilon$ networks both indicate that the datasets comprise a single connected component (Fig. S14). Taken together, these findings suggest that the manifold hypothesis actually does not hold for scRNA-seq data, calling into question the widespread application of that hypothesis in the literature (47–49).

MNIST Digits Data

Since we did not find the expected manifold structure in our analyses of various scRNA-seq data, we sought to verify DeepAtlas on another higher-dimensional dataset where a lower-dimensional manifold is thought to exist: the popular dataset of handwritten digits from the Modified National Institute of Standards and Technology (MNIST) database (50). This is a set of 64-dimensional images of handwritten digits 0 through 9, where each value represents the intensity of a pixel of an 8x8 image (Fig. 6A).

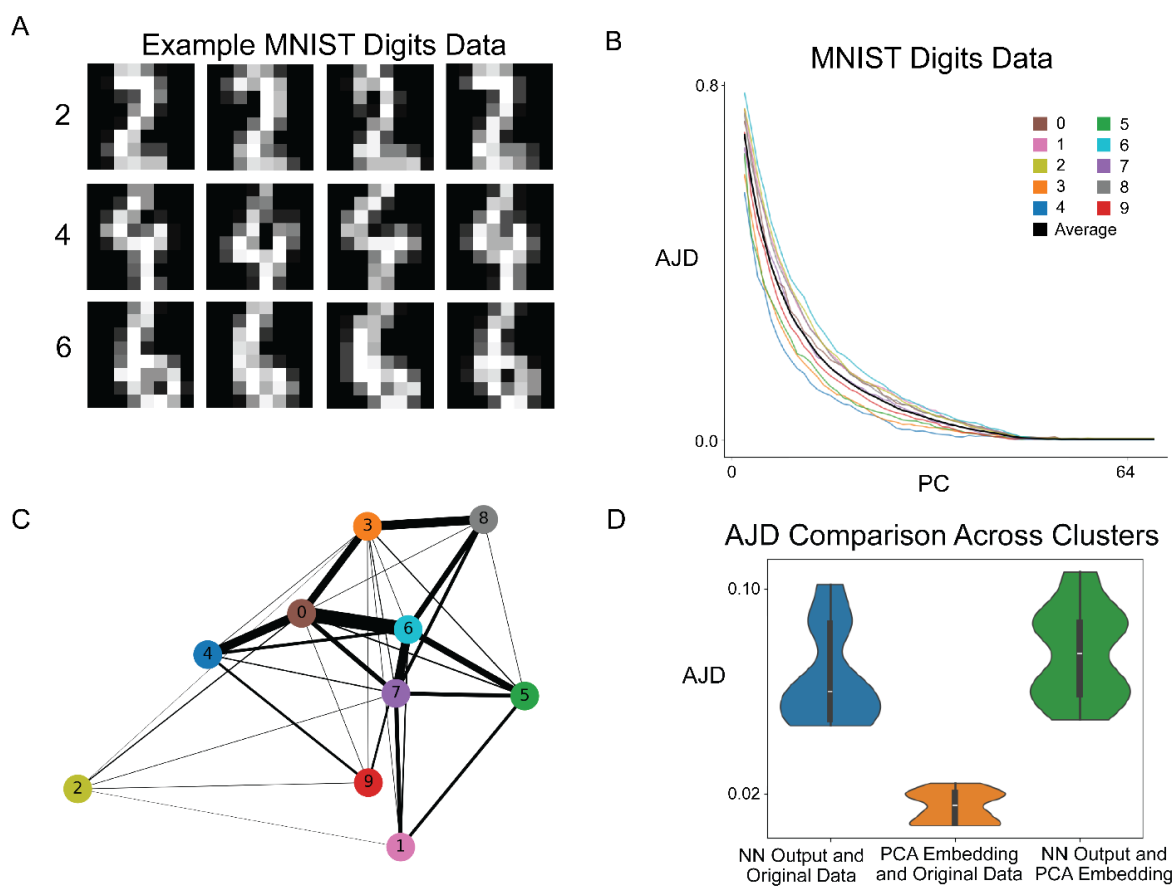

**Fig. 6. MNIST Digits Show Manifold Structure.** (**A**) Example images from the MNIST Digits dataset, which consists of 8x8 pixels representing handwritten images of digits 0-9. (**B**) PC vs. AJD plot of the MNIST data with $k = 10$ clusters, indicating a lower dimensional manifold around dimension 40. (**C**) A graph representation of the MNIST dataset. Each node corresponds to one of the $k = 10$ clusters and edges are placed between clusters that have shared nearest neighbors. The clusters are numbered according to which number in the MNIST set makes up the majority of that cluster. The colors here match panel B. The width of each edge is proportional to the number of transition points between the clusters. (**D**) AJD comparison after applying DeepAtlas and creating a neural network model.

We applied DeepAtlas to this data using $k = 10$ for $k$-means clustering, as we expect each of the ten digits to cluster together. The plot of PC vs. AJD for each local neighborhood does indeed indicate the existence of a lower-dimensional manifold (Fig. 6B). The lines representing each cluster are close together and show similar behavior across PC dimensions. Further, all clusters show little to no distortion in PC dimensions above 40. This, combined with the fact that the data consists of a single connected component (Fig. 6C and Fig. S15), suggests that the MNIST data is drawn from a 40-dimensional manifold. We then learned the functions to map from these local 40D PCA embeddings to the 64D original data. Since we are training on image data, we found better performance when using convolutional layers in our neural network. The trained model results in low AJD values, which means there is little distortion between the model output and the true data (Fig. 6D). This demonstrates that DeepAtlas can be trained to construct a differentiable model of real datasets in relatively high dimensions.

Model Applications

The fact that DeepAtlas learns a model of the manifold enables a number of interesting applications. For one, unlike some manifold learning tools, it is very straightforward to apply DeepAtlas to new data that might be experimentally generated for the same system. To map any new data point(s) to their local lower-dimensional representations in the model, all one has to do is assign each point to the appropriate cluster and then apply the PCA transformation for that cluster to represent that point in the local embedding. In the current implementation, assigning a point to a cluster is straightforward: one can do so simply by calculating the distance between the point and all the "$k$ means" and choosing the cluster that is closest. So the nature of DeepAtlas makes it very extensible to model new data points.

Another example application of DeepAtlas is to sample points from the low-dimensional representation and then use the charts to map these new points back to the original manifold on which the model was trained. This effectively allows us to computationally sample *new* points on the same manifold without having to perform any further experiments. The idea here is that we first sample points in the lower-dimensional representation for each local cluster, and then use the neural networks to map these new points back into the high-dimensional space to generate new examples drawn from the same manifold. To sample a new point from the lower-dimensional representation, we first pick one of the original data points at random. We then sample "nearby" this original point by sampling from a ball of radius $r$ surrounding the point, where $r$ is the distance to one of the point's nearest neighbors (Fig. 7A). We can then use the appropriate neural network in the atlas to map this new point back to the original manifold.

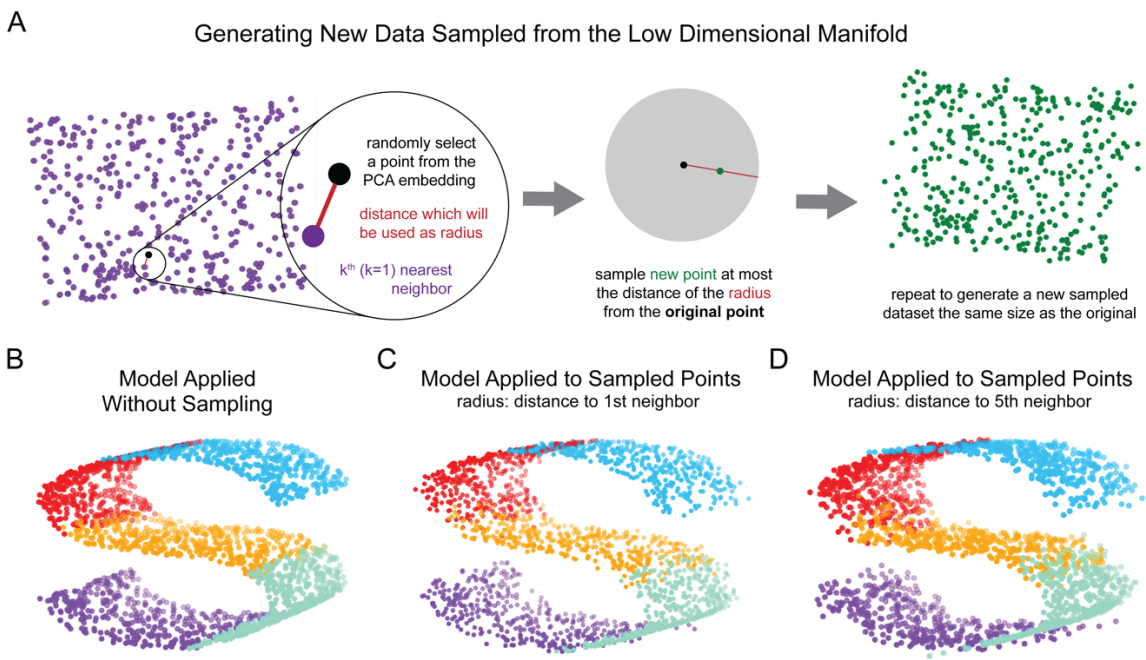

Fig. 7. Using DeepAtlas Generatively. (A) Schematic of the process used to generate new data sampled from a manifold. (B) The model applied to the PCA embedding directly without sampling. (C) The model applied to the sampled points generated with the radius of the 1st nearest neighbor. (D) The model applied to the sampled points generated with the radius of the 5th nearest neighbor.

To demonstrate that DeepAtlas can be used generatively in this way, we revisited the S-curve dataset. We used the procedure above to sample exactly as many data points from each cluster as we had in the original dataset (Fig. 7A). As expected, when we apply DeepAtlas to the original points, we recover the S-curve (Fig. 7B). When applied to a new dataset of points sampled from the PCA embedding using a radius of the $1^{st}$ and $5^{th}$ nearest neighbor, the model was again able to generate the S-curve structure (Figs. 7C, 7D). This was also the case with larger radius values, though we do see more error as the radius increases (Fig. S16). This experiment demonstrates that DeepAtlas can be used to generate a new dataset that lies on the same lower dimensional manifold as the training data by sampling from the local embeddings.

**Discussion**

DeepAtlas takes a novel approach to manifold learning compared to previous tools. First, it provides a way for researchers to explore the assumption that their data is actually drawn from a lower-dimensional manifold. By examining the AJD at each PC dimension for different local neighborhoods in the data, one can determine not only whether a lower dimensional manifold exists, but also determine the dimensionality of that manifold. The analyses presented in Figs. 3, 5 and 6 thus provide a valuable empirical approach to exploring the manifold hypothesis for any given dataset. Second, DeepAtlas focuses on finding local embeddings of the data that minimize topological distortion. All other "manifold learning" algorithms of which we are aware attempt to generate a global embedding. Finally, DeepAtlas trains a neural network, yielding a set of differentiable functions that, in conjunction with PCA, approximates a set of invertible charts that can map back-and-forth between high dimensional data and the lower dimensional local representations. This model *directly represents* the standard mathematical definition of a (differentiable) manifold, and can be used generatively to create a new dataset with points that are sampled from the manifold itself. In these key ways, DeepAtlas addresses gaps in the existing standard practice, improves upon current methods, and moves the field of manifold learning in a new direction.

We have validated DeepAtlas approach using a variety of datasets. Several of these are designed to be manifolds, such as the S-curve and hyperspheres. In all of these test cases, DeepAtlas succeeded in: 1) identifying that the data was actually drawn from a manifold; 2) finding low-distortion local embeddings at the correct local dimension; and 3) using simple deep neural networks to train a set of differentiable functions that can map from the low-dimensional local embeddings to the original data with very low topological distortion. This stands in stark contrast to the performance of existing manifold learning tools, which cannot recover low-distortion global embeddings even for hyperspheres (24).

Interestingly, although the real-world data we explored is commonly assumed to be drawn from a lower dimensional manifold (30), most of the real-world datasets we considered here were not consistent with the manifold hypothesis. In particular, many of the machine learning test datasets and all of the single cell data we considered did not have a consistent local embedding dimension across various local neighborhoods. In the datasets that did exhibit evidence of being drawn from a manifold, DeepAtlas' local embedding approach resulted in an overall lower AJD than all the traditional manifold learning algorithms we tested. From there, DeepAtlas was able to train a neural network to approximate the inverse of the chart for each local cluster. This model can be used to determine the location of a new point or can even be used generatively, as seen on the S-curve dataset.

While DeepAtlas is promising, the approach presented here has clear limitations. The diagnostic process of determining whether the data is likely drawn from a low-dimensional manifold, and identifying the local dimension of that manifold, is currently done manually through visual inspection. Because of this, there may be some subjectivity in a researcher's interpretation. Additionally, the neural network training process may require changes to the layer structure and other parameters to achieve an effective model. For instance, our neural network model of the MNIST data used a convolutional layer whereas the other models trained used simple fully connected layers. As such, researchers may see different results depending on their decisions regarding certain parameters. While the approach is thus an iterative one, requiring extensive guidance from the user, DeepAtlas does provide extensive diagnostics that allow researchers to evaluate how well their model represents the data at any given step. In future work, we hope to improve upon the robustness of DeepAtlas and reduce the need for manual inspection and input.

Another limitation of this approach is the reliance on specific algorithms, notably $k$-means clustering to identify local neighborhoods and PCA to generate local embeddings. In our experience, none of the non-linear dimensionality reduction tools broadly employed in the field generate lower distortion local embeddings than PCA, but improvements in such embedding tools could greatly enhance the performance of DeepAtlas. Similarly, $k$-means clustering was used here for convenience and for its utility in identifying clusters that correspond to local areas in the dataset (i.e. clusters of points that are localized in the same region of the original high-dimensional space). The design of DeepAtlas is highly modular, allowing for different clustering and dimensionality reduction tools to be easily substituted based on the users' preference. Importantly, the AJD curve and the cluster size distribution can be used to evaluate how such choices ultimately perform. In future work, we aim to explore more thoroughly different clustering and dimensionality-reduction tools on a range of real and synthetic data in order to provide potential users with a more comprehensive picture of how various tools perform in this context. Finally, future work will be needed to further explore DeepAtlas' generative capabilities, and to investigate new applications for the output model, including leveraging the model for direct application of ideas from differential geometry and topological data analysis (51, 52). While future development of DeepAtlas is clearly necessary, we hope that this tool motivates and enables deeper exploration of manifolds in big data across a variety of fields.

## Materials and Methods

Clustering
DeepAtlas begins by first partitioning the data into local neighborhoods. These neighborhoods will then be embedded into the lower dimension and can then be stitched back together. Through this process, DeepAtlas takes a local approach which allows for less topological distortion. Here, we use $k$-means clustering to generate the local neighborhoods, which results in $k$ clusters of points grouped by distance to their centroid. This method aligns with our intuition that the clusters should be composed of points which are local neighbors. DeepAtlas is designed to be modular, so the user can choose to substitute a different clustering algorithm if desired.

The $k$-means clustering algorithm requires the user to choose a value of $k$ clusters to generate. The choice of this parameter can affect DeepAtlas' performance. One easily illustrated example of this is when $k$-means is applied to the Swiss roll dataset. When a smaller value of $k$ is chosen, the clusters may contain points from different sections of the unraveled plane. When a sufficiently large value of $k$ is chosen, each cluster contains only points which are next to each other when the manifold is unraveled. This allows for more accuracy when embedding each local neighborhood and then subsequently stitching them back together into the 2D manifold. With higher dimensional data, it is more difficult to determine the optimal value of $k$ to use. We tested varying this parameter in our experiments to ensure that the downstream results were not affected. Generally, we proceed with a value of $k$ such that each cluster contains at least 50 points.

Transition Points
Transition points are the sets of points which connect the clusters. To determine them, we first take the original dataset and determine the $l$ nearest neighbors of each point. A point is in the transition region if one of its $l$ nearest neighbors is in a different cluster from the one to which it is assigned. While this $l$ is a free parameter of DeepAtlas, here we focus on a value of 10, which works well for the example datasets we considered. We then add the transition points to each of our $k$ clusters in a simple way: if a point that is not in a cluster is one of the $l$ nearest neighbors of a point in that cluster, we add that point to the cluster. This results in a set of points that are present in each neighboring cluster and when overlayed, connect the clusters together.

Quantifying Distortion
The metric we use to quantify distortion here is the Average Jaccard Distance (AJD). This measures the dissimilarity between a dataset and its embedding by comparing the nearest neighbors of each point. If all points have the same set of nearest neighbors in both the original data and the embedded data, the AJD will be 0. Conversely, if all points have a completely different set of neighbors in the embedded data than in the original data, the AJD will be 1. The Jaccard distance for each point can be calculated as:

$$J_i = \frac{|A_i \cup B_i| - |A_i \cap B_i|}{|A_i \cup B_i|} \quad (1)$$

where $A_i$ is the set of $h$ nearest neighbors of a given point $i$ in the original data, $B_i$ is the set of $h$ nearest neighbors of $i$ in the lower-dimensional representation, and $|X|$ represents the cardinality of the set $X$. To get the AJD, we then take the average of the Jaccard distances of every point in the dataset:

$$AJD = \frac{1}{N}\sum_i J_i \qquad (2)$$

where $N$ is the total number of points in the dataset.

PC vs. AJD Plots
DeepAtlas outputs a plot of the AJD between the original dataset and the embedding into every lower dimension using PCA. This PC vs. AJD plot helps the user determine whether a lower dimensional manifold exists, and if so, what the dimensionality of the manifold is. Each cluster is represented by a line on the plot. If the original data is drawn from a lower dimensional manifold, the clusters' lines will all follow a similar curve, approaching 0 as the number of PCs approaches the dimensionality of the manifold. For example, in the simulated hypersphere data, the clusters' lines are so similar that they appear on top of each other (Fig. 3A in the main text). Each cluster has a similar level of distortion at each PC, indicating that the data in each cluster is from the same manifold. The AJD is 0 at higher dimensions, and begins to rise at the minimal embedding dimension of the hypersphere. As the number of PCs decreases further, the AJD increases significantly.

However, if the clusters' lines are not aligned, this suggests that there is not a common lower-dimensional manifold across clusters. For example, in the Boston Housing data, the clusters reach 0.1 AJD at varying PCs from 3 to 9. This wide range suggests that the clusters are not on the same lower-dimensional manifold. Visual inspection of these plots can be used to adjust hyperparameters in the data, such as the value of $k$ used for $k$-means clustering.

Local Embedding
Once it has been determined that a lower-dimensional manifold exists and what its dimension is, we can proceed with embedding the data into that dimension. Each cluster is embedded separately and, since we have included transition points in each, the embedded clusters can then be re-aligned or "stitched" back together in the lower-dimensional space. This local approach results in less distortion overall as measured by AJD. Here we use PCA to embed each cluster into the lower dimension, as our previous work has found it to be the only method that could reliably reduce high-dimensional data with low distortion (24). Again, the modularity of DeepAtlas allows for the user to instead apply any other dimensionality reduction method of their choice.

Generating a true embedding of the S-curve
We generated all of our S-curve datasets using the "make_s_curve" function in the scikit-learn library in Python (29). The 3-dimensional S curve is generated by first drawing two random numbers. The first, $t$, is drawn from a uniform distribution of real numbers on the interval $\left[-\frac{3}{2}\pi, \frac{3}{2}\pi\right]$; the second, $h$, is drawn from a uniform distribution on $[0, 2]$. For any given value of these two random numbers, which we represent as a 2-vector $(t, h)$ we can generate a corresponding 3-dimensional point using the following function:

$$f\big((t,h)\big) = (\sin(t), h, \text{sign}(t) \cdot \cos(t-1)) \qquad (3)$$

where we have written out the three components of the output, and $\text{sign}(t)$ is the sign function, returning $-1$ if $t$ is negative, $+1$ if $t$ is positive, and 0 otherwise. Evidently, the function $f^{-1}$ is a global 2D embedding of the 3D S-curve manifold generated by this approach. In Fig. 1 we use $(t, h)$ for the "true" embedding of the corresponding S-curve dataset.

Neural Network Charts as Diffeomorphisms
As mentioned in the main text, if we claim to generate a differentiable atlas of a manifold, we need to show that our approach approximates charts that are diffeomorphisms. To do so, here we set up a slightly more formal discussion of DeepAtlas' framework and then show that the neural networks we train are guaranteed to satisfy the axioms needed to exhibit an atlas of a differentiable manifold.

In the main text, we stated briefly the requirements for a set $M$ to be considered a differentiable manifold. To restate this, such a set is a differentiable manifold if and only if:

1. There exists a covering of open subsets of $M$, $\{U_i\}$, such that each $U_i$ is diffeomorphic to an open subset $V_i \subset \mathbb{R}^n$. For any member of this covering $U_i$ we refer to the corresponding diffeomorphism as $\phi_i: U_i \to V_i \subset \mathbb{R}^n$.
2. If two sets $U_i$ and $U_j$ in this covering have a non-trivial intersection, i.e. $U_i \cap U_j \neq \emptyset$, then $\phi_j \circ \phi_i^{-1}$ is (smoothly) differentiable on $\phi_i(U_i \cap U_j)$.

The above is a standard definition of a differentiable manifold, and further (and more formal) treatments can be found in textbooks on the subject (27, 28).

First, we note that DeepAtlas as described in the main text is trained on finite sets of points; each point in the data set is a vector in a real-valued vector space $\mathbb{R}^m$. Call this set of points $D$, and say that $D$ is a finite sample of points from the manifold $M$. Since we don't know this manifold *a priori*, $M$ is just taken to be an appropriate collection of neighborhoods around the points in $D$. Similarly, to construct a covering of $M$, we use the $k$-Nearest Neighbors clustering algorithm to first generate a partition of $D$ with $k$ subsets. We then add the transition points between neighboring clusters to those relevant clusters as described above, generating a set of subsets of $D$; call this set of subsets $\{C_i\}$. Note each element of this subset is finite and thus closed; for the formal purpose of developing the differentiable structure here, we associate with each $C_i$ an collection of open neighborhoods "around" the points in $C_i$ to form a corresponding open subset $U_i$ in such a way that the collection $\{U_i\}$ forms a covering of $M$. Note that we do not do this explicitly; this is just an implicit step in our algorithm to satisfy the mathematical definition above.

Now we have our covering of open subsets, and we have to generate a set of diffeomorphisms $\phi_i: U_i \to V_i \subset \mathbb{R}^n$ that will represent our charts. In our case, we have $U_i \subset \mathbb{R}^m$, so every input to our function will just be a point in an $m$-dimensional real vector space. DeepAtlas approximates each $\phi_i$ using two separate pieces. To define the "forward" direction, from $U_i$ to $\mathbb{R}^n$ we use PCA. Note that PCA is a linear function and is thus smooth. We then train a deep neural network to approximate the inverse of these functions ($\phi_i^{-1}$). Certain aspects of the architecture of these networks vary depending on the application area, but all these networks have the same basic mathematical structure. For each node at any given layer in the network, we have a set of incoming "links," see Fig. 4A. Denote the output of some layer $l - 1$ as $x_{l-1}$; the first step is to compute an intermediate value $z_l = W_l x_{l-1}$, where $W_l$ is the weight matrix at that layer. Note that $z_l$ is a vector. The next step is to produce the output of the layer, $x_l$, by applying a non-linear activation function to *each element* of the vector. We used the activation function $a(z) = \tanh(z)$, the hyperbolic tangent, so $x_l = \tanh(z_l)$. Note that each of these steps involves the use of a smooth function. Since each layer is a composition of smooth functions, and the entire function defined by the neural network is just a composition of smooth functions and thus is, itself, smooth.

As a result, we now have two smooth functions to approximate each chart; the linear PCA step to map from $\mathbb{R}^m$ to $\mathbb{R}^n$ (i.e. to reduce dimensionality) and the neural network to map back to $\mathbb{R}^m$ from $\mathbb{R}^n$. We use these two functions separately to approximate the chart, and since these two directions are both smooth, this allows us to approximate the diffeomorphisms in the atlas.

Network Training

The neural network used by DeepAtlas is modular. The user should adjust the parameters to achieve a model that best fits their data. For our experiments here, we use a simple fully connected network implemented through dense layers using the Tensorflow and Keras libraries in Python3. The networks are trained with 10 layers for 10,000 epochs. We use the hyperbolic tangent function $a(x) \equiv \tanh(x)$ as the activation function. The network is trained using stochastic gradient descent with Adam optimization. It is validated using mean squared error as the loss during training. Once the training is complete, we further evaluate the model by calculating the AJD of the network's output compared with both the original data and the PCA embedding. The only network with a different structure is the one used for the MNIST data, where we instead included some convolutional layers.

Cross Validation

To evaluate the performance of our neural network, we perform 10-fold cross validation. The data is split into 10 "folds" or partitions. The network is then trained with one fold left out to be used for validation each time. This demonstrates the model's robustness, showing that the resulting low loss is repeatable and not due to overfitting.

Single Cell RNA-Seq Transformations

Raw single cell RNA-sequencing data refers to data which has not undergone any preprocessing. Following the typical standard practice, we apply 3 steps of preprocessing and test DeepAtlas at each stage. The first is to select the highly varying genes, proceeding with only this subset of the data. Next, the data is scaled using counts-per-million normalization. Last, the dataset undergoes log transformation. Typically, scRNA-seq data is analyzed in this last stage after all preprocessing steps are complete. However, we were interested in seeing whether a manifold existed in any stage and thus applied DeepAtlas after each preprocessing step.

Simulated scRNA-Seq Data

We tested DeepAtlas on simulated scRNA-Sequencing data that was made using a package called scDesign[42]. This package uses real scRNA-seq data to generate a synthetic dataset. We applied it with default parameters to scRNA-seq cell line data in order to obtain a simulated scRNA-seq dataset to use.

Generative Applications

Because DeepAtlas' neural network is designed to be bidirectional, we can also apply the learned functions to data drawn from the same lower-dimensional manifold to obtain its position in the original high-dimensional space. We demonstrate this on the S-curve data by first generating new data sampled from the local 2D representations of the subsections of the manifold. The new data is generated by selecting a point from the PCA embedding of the real data at random and sampling from a ball surrounding that point, using a radius based on nearest neighbors. Our example shows the results when the data is sampled with a radius of both the 1$^{st}$ nearest neighbor and the 5$^{th}$ nearest neighbor. This sampling process is repeated until the generated dataset is of the same size as the original dataset (although one can in principle sample any number of desired points). We can then pass the generated dataset into the functions learned by DeepAtlas' neural network. This yields the high-dimensional space representation of the new dataset sampled from the same low-dimensional manifold. As shown in the main text, this successfully allowed us to generate S-curve shapes from data sampled from the lower-dimensional space.

Statistical Analysis
Cross-validation calculations were carried out using 10-fold cross validation throughout the study, as indicated. DeepAtlas does not entail the use of other statistical approaches at present.

# Acknowledgments

**Funding:**
National Institutes of Health grant R01GM143378 (EJD)
National Heart Lung and Blood Institute grant T32HL139450 (SJH)

**Author contributions:**
    Conceptualization: SJH, TMK, EJD
    Methodology: SJH, TH, TMK, EJD
    Investigation: SJH, TH, EJD
    Visualization: SJH, EJD
    Supervision: EJD
    Writing—original draft: SJH, EJD
    Writing—review & editing: SJH, TH, TMK, EJD

**Competing interests:** The authors declare that they have no competing interests.

**Data and materials availability:**
All Code and example datasets can be found at [https://github.com/DeedsLab/DeepAtlas](https://github.com/DeedsLab/DeepAtlas).

# Supplementary Materials for

## DeepAtlas: a tool for effective manifold learning

Serena J. Hughes *et al.*

Corresponding author: Eric J. Deeds, Email: deeds@ucla.edu

**This PDF file includes:**

Figs. S1 to S16

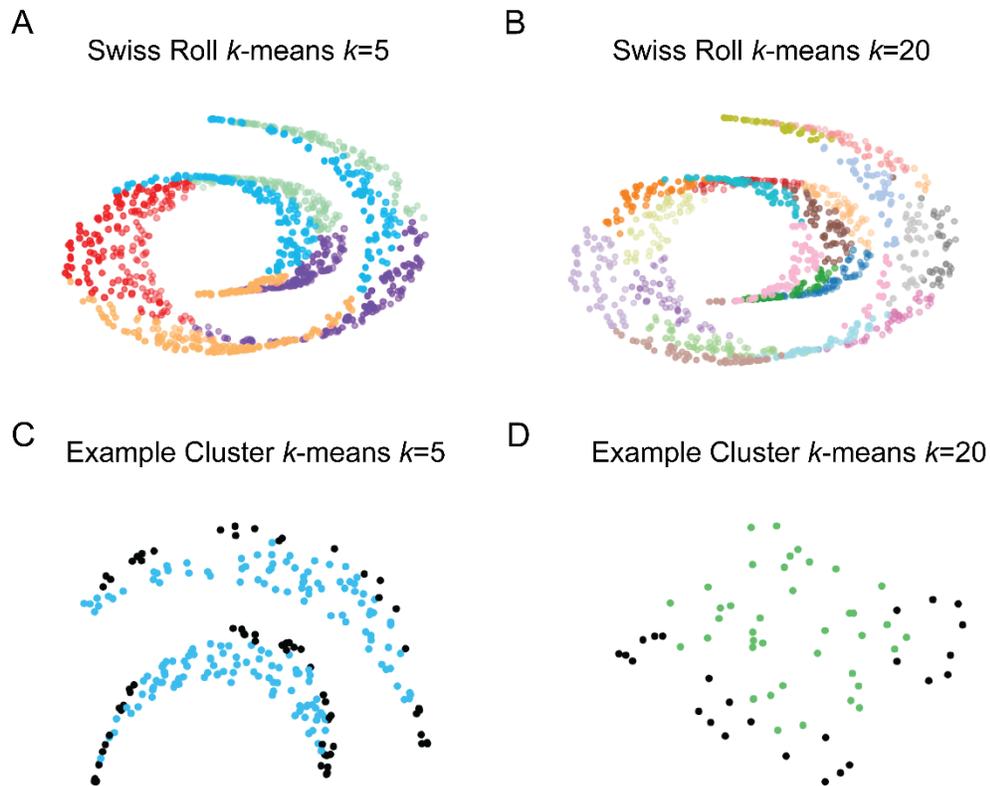

**Fig. S1.**
A) Swiss roll dataset in 3D with points colored based on $k$-means clustering using $k = 5$. B) Swiss roll dataset in 3D with points colored based on $k$-means clustering using $k = 20$. C) Example of one cluster from the Swiss roll data shown in S1A where points that are not nearby in the original dataset have been clustered together. D) Example cluster from the Swiss roll data shown in S1B. While some clusters still contain points that aren't neighbors in the 3D data, using more clusters reduces the occurrence.

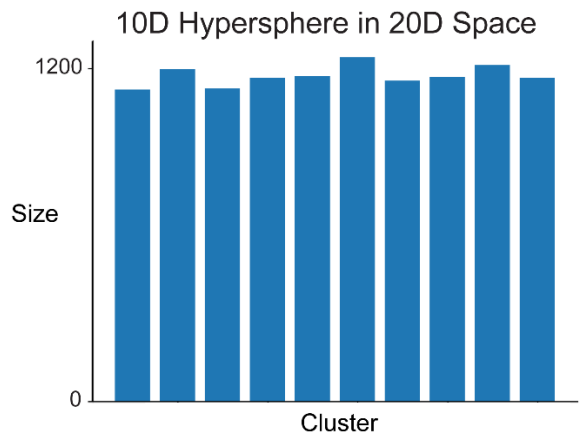

**Fig. S2.**
Cluster composition of the $k$-means clusters of the 10-dimensional hypersphere data embedded in 20-dimensional space, using $k = 10$.

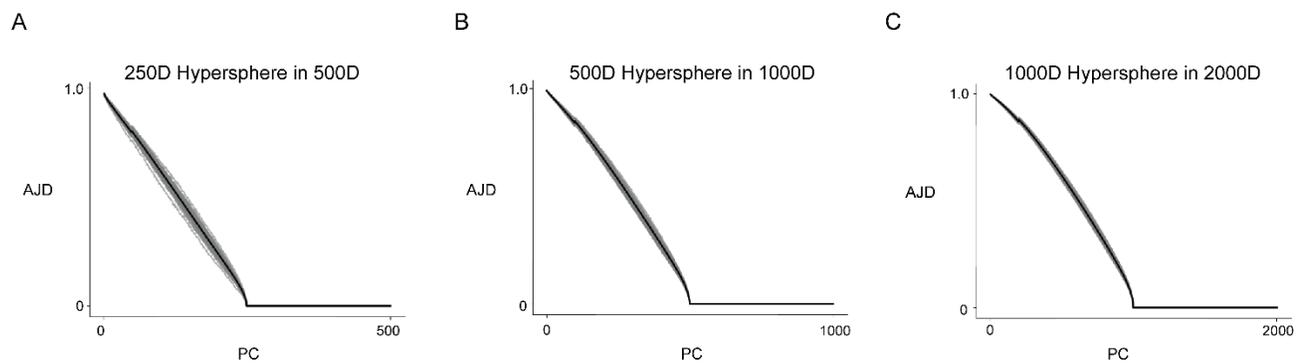

**Fig. S3.** PC vs. AJD for higher dimensional hyperspheres. A) 250-dimensional hypersphere of 5,000 points embedded in 500-dimensional space. B) 500-dimensional hypersphere of 10,000 points in 1000D space. C) 1000-dimensional hypersphere of 20,000 points in 2000-dimensional space.

**Fig. S4.**

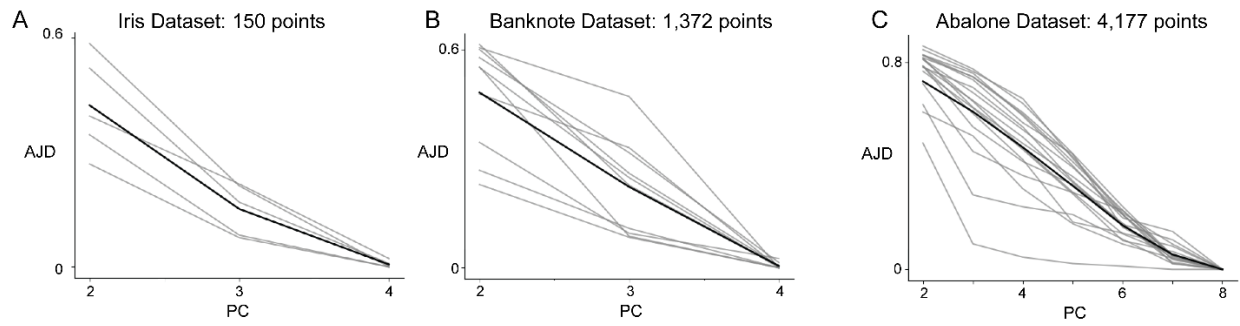

PC vs. AJD plots of other standard test machine learning datasets which do not indicate a lower dimensional manifold. A) Iris dataset with 4 dimensions and 150 points. B) Banknote dataset with 4 dimensions and 1,372 points. C) Abalone dataset with 8 dimensions and 4,177 points.

## The Union of a Sphere and a Circle

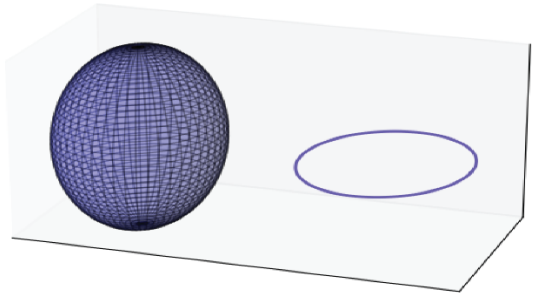

**Fig. S5.**
One dataset consisting of both a sphere and a circle, each of which is its own manifold.

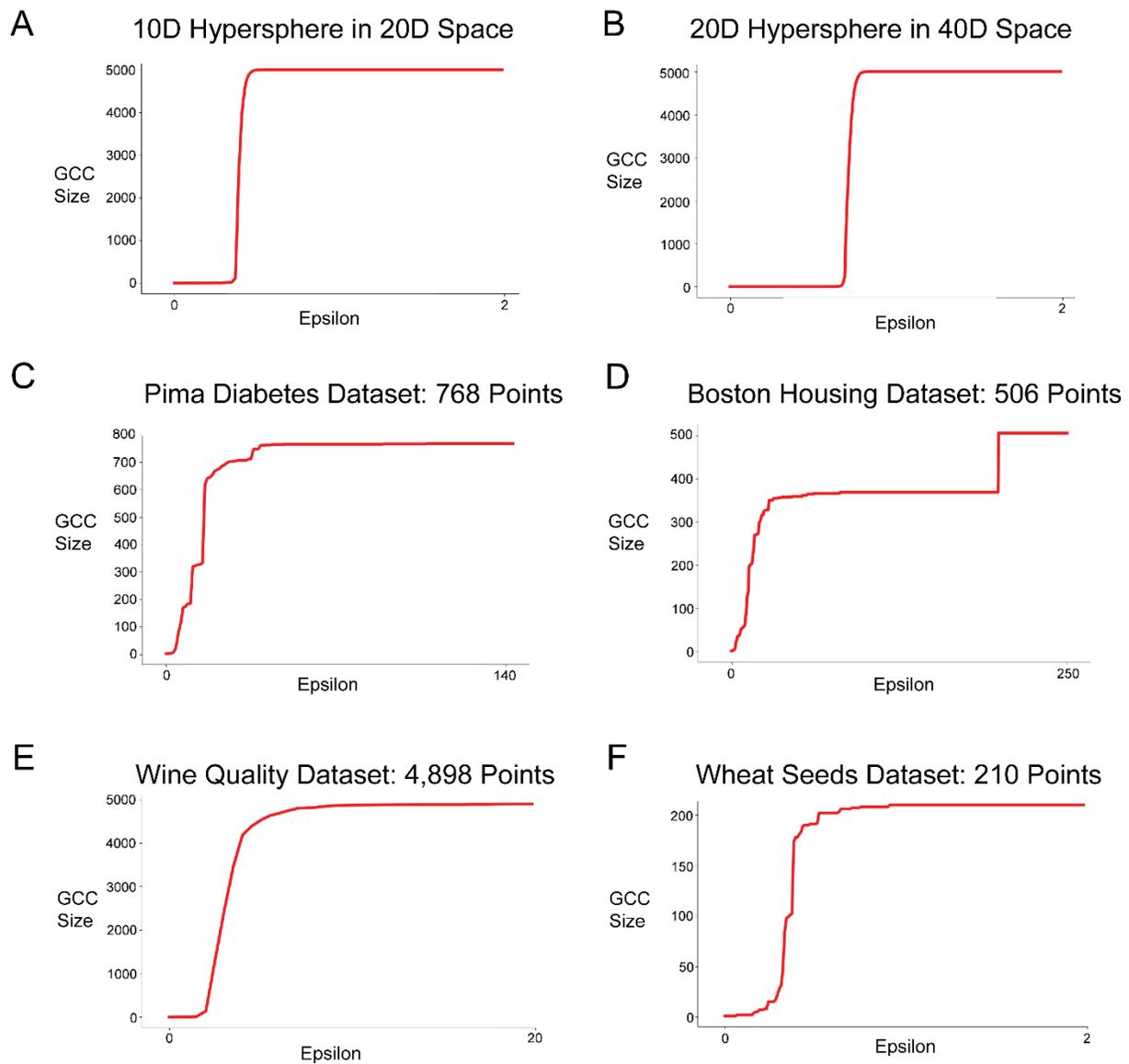

**Fig. S6.**
Epsilon plotted against the size of the giant connected component for each dataset. Smooth curves indicate the data is composed of one connected component while step behavior shows multiple components that are being joined together.

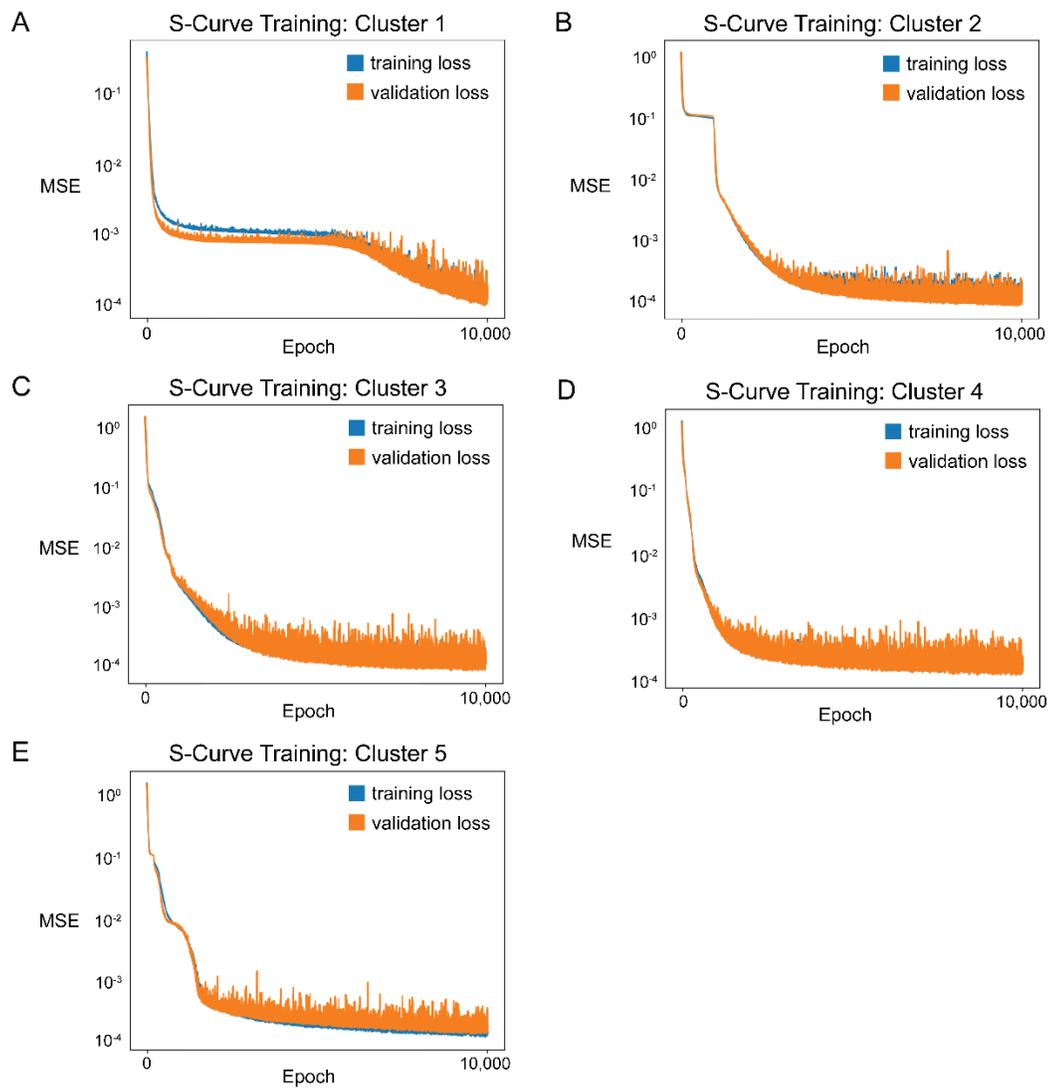

**Fig. S7.**
Training results of DeepAtlas' neural network on each of the S-curve clusters, generated using $k$-means with $k = 5$.

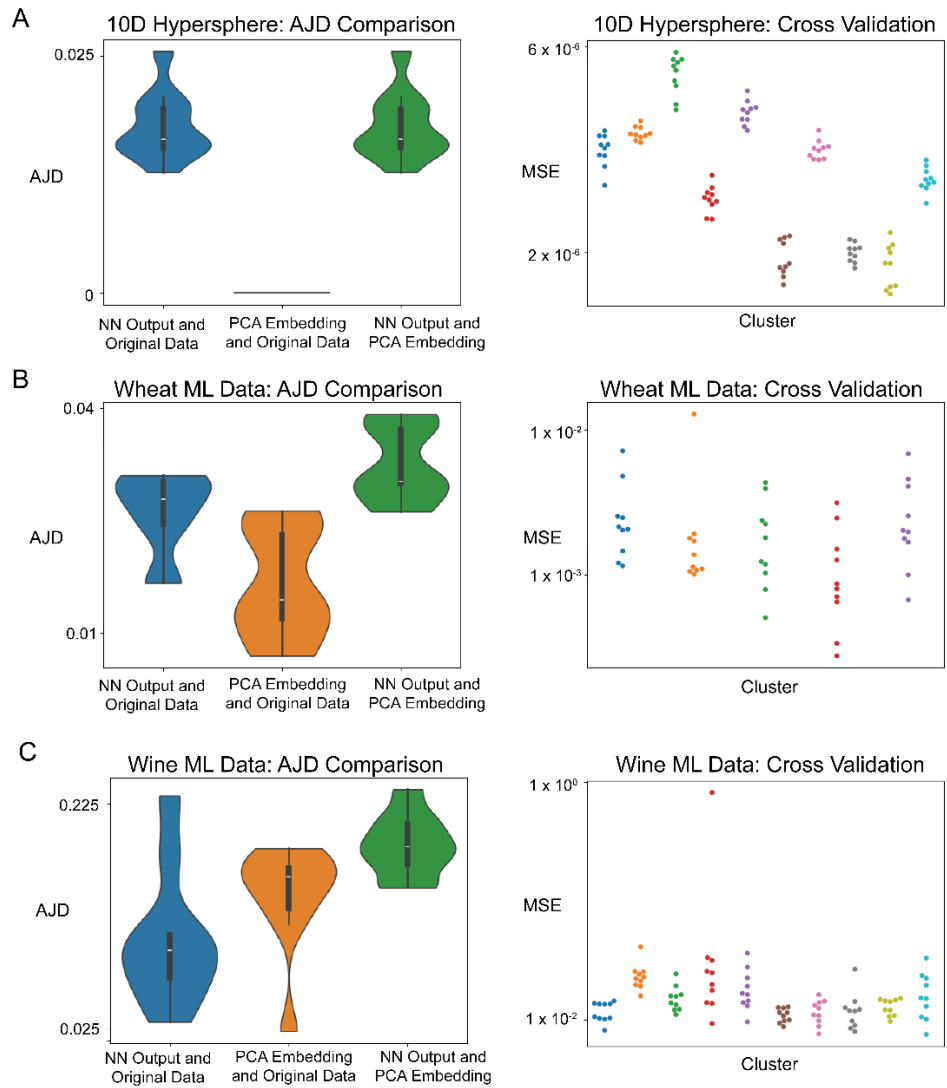

**Fig. S8.**
AJD comparison and 10-fold cross validation results for DeepAtlas applied to other datasets which indicated a lower dimensional manifold. A) 10D Hypersphere embedded into 20D space, with a lower dimension at 10D. B) Wheat machine learning test data with a lower dimension at 4D. C) Wine machine learning test data with a lower dimension at 10D.

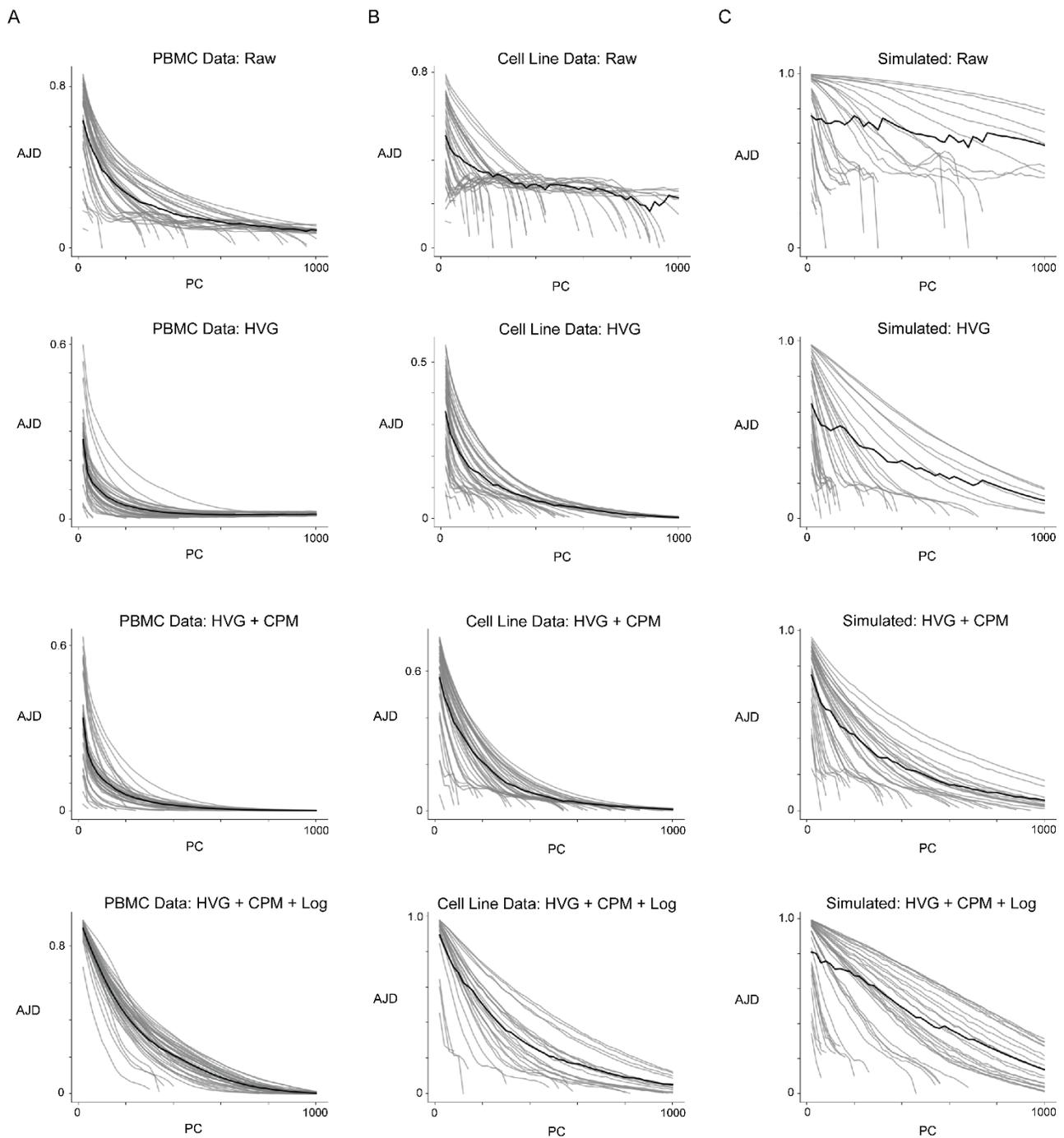

**Fig. S9.**
PC vs. AJD plots for additional single cell RNA Sequencing datasets explored, across all preprocessing steps. A) The full PBMC data. B) Cell Line Data. C) Simulated data generated using sc-Design, based on the cell line data.

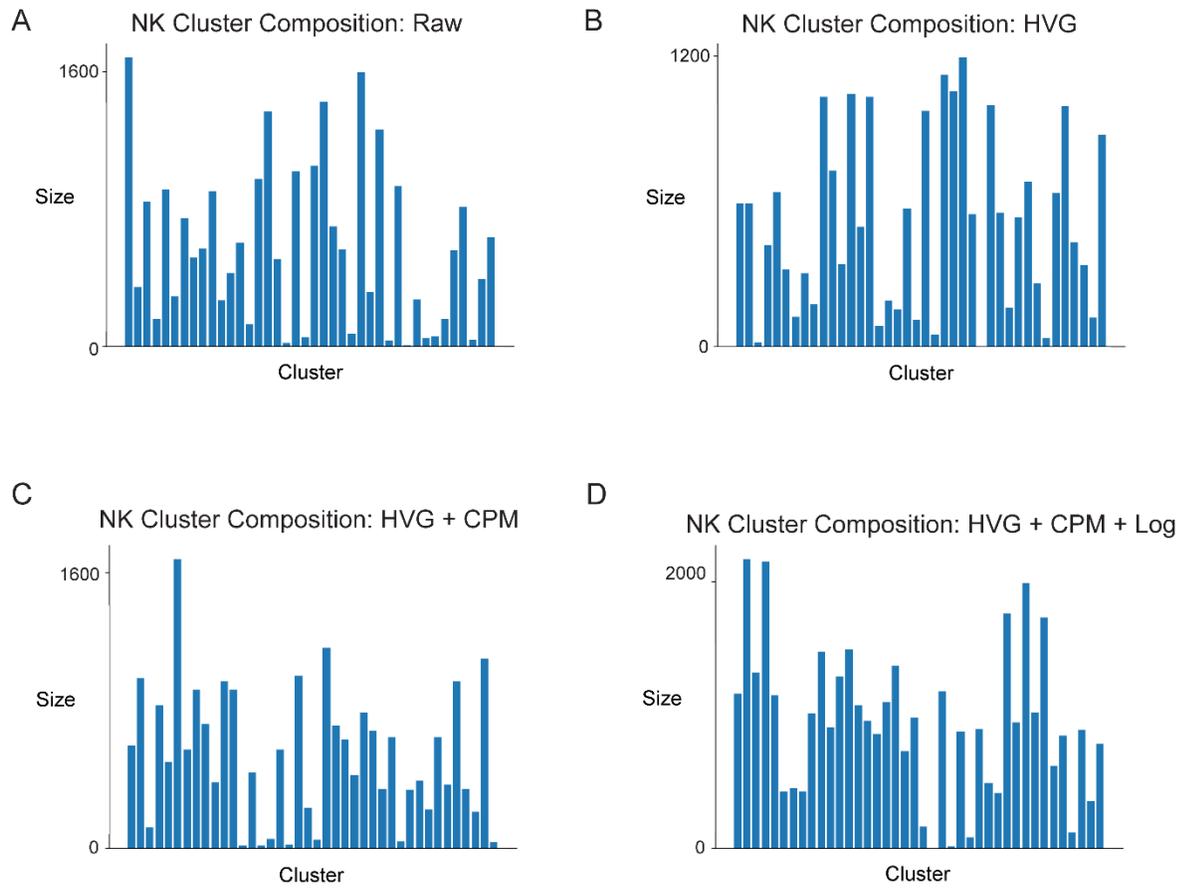

**Fig. S10.**
NK cell scRNA-seq data cluster sizes across preprocessing stages. The trend of inconsistent sizing occurs in each.

**Fig. S11.**

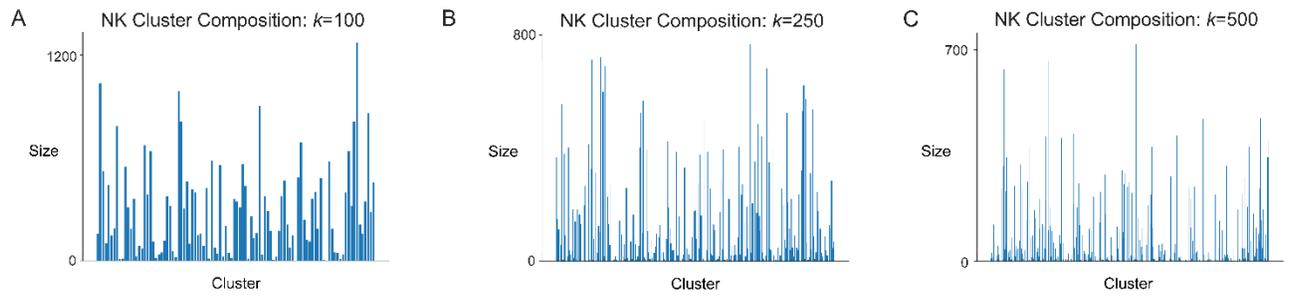

Raw NK cell scRNA-seq data cluster sizes when using different values of $k$ for $k$-means. See S10 for $k = 40$.

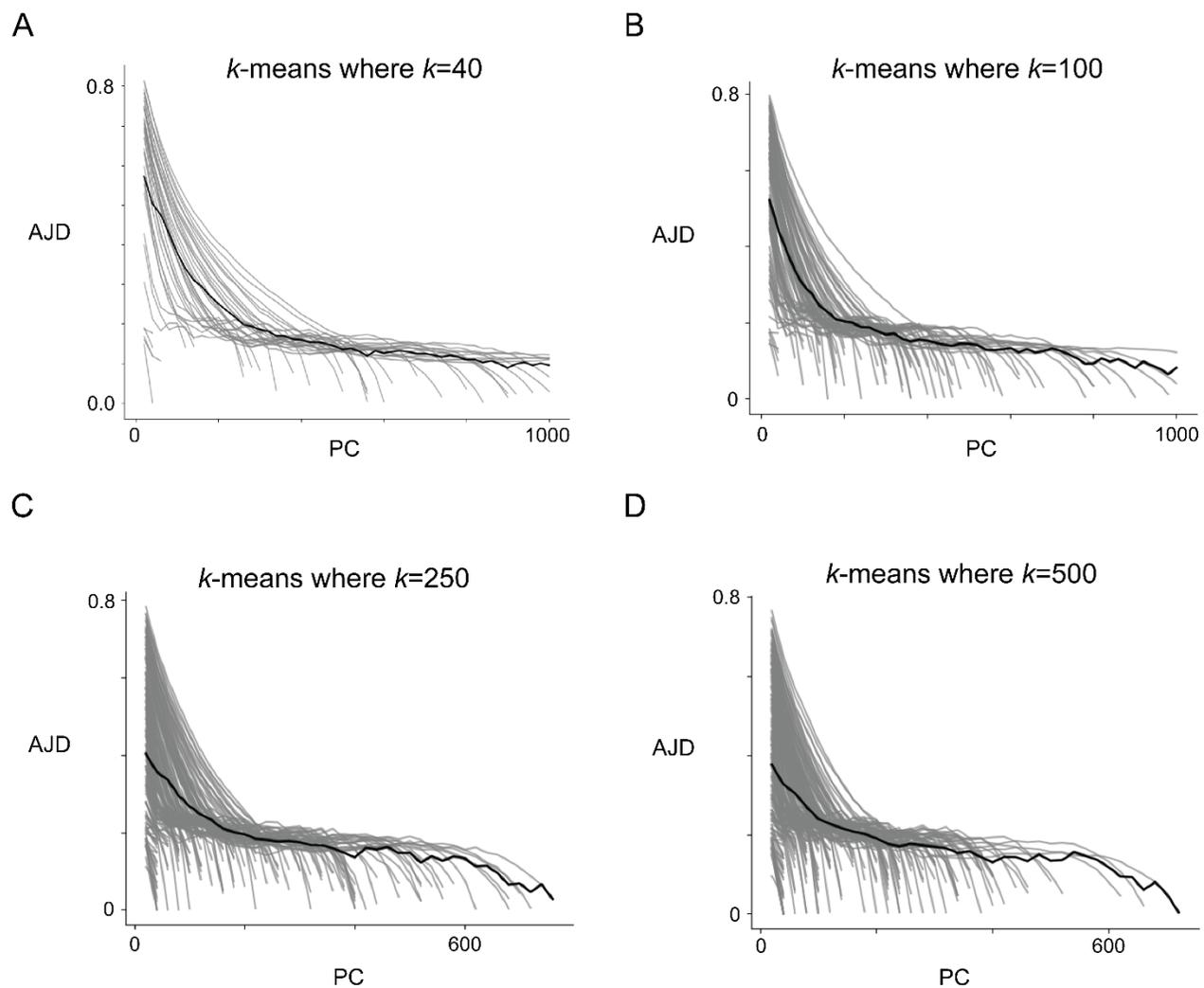

**Fig. S12.**
PC vs. AJD plots for the raw NK cell scRNA-seq data when using different values of $k$ for the $k$-means. A) $k = 40$, as shown in the main text. B) $k = 100$. C) $k = 250$. D) $k = 500$.

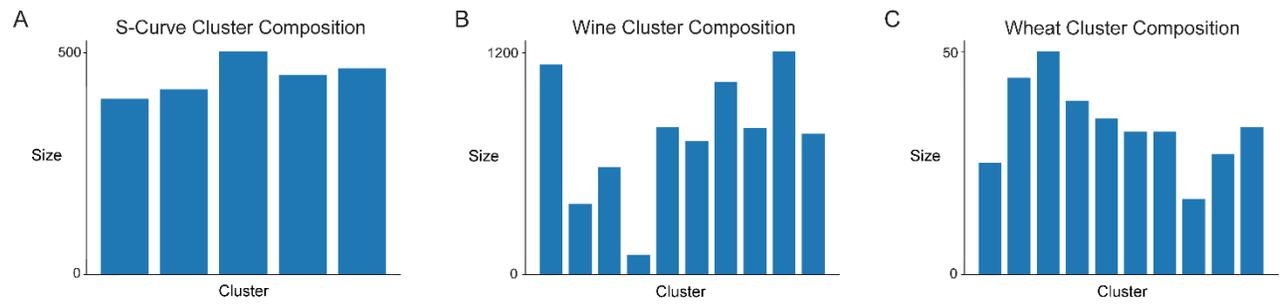

**Fig. S13.**
Cluster sizes when $k$-means is applied to other datasets which did indicate a lower dimensional manifold. A) S-Curve data with $k = 5$. B) Wine machine learning data with $k = 10$. C) Wheat machine learning data with $k = 10$.

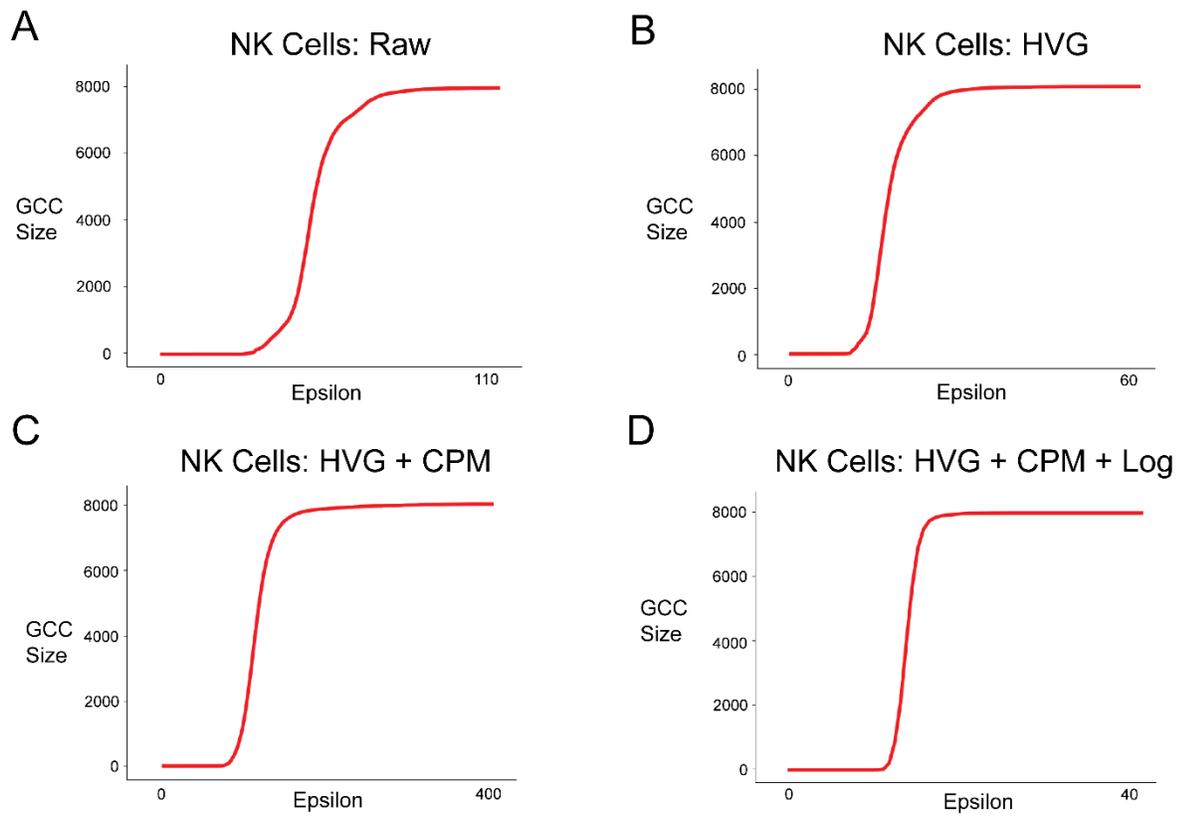

**Fig. S14.**
Epsilon plotted against the giant connected component size for the NK cell scRNA-seq data at each preprocessing stage.

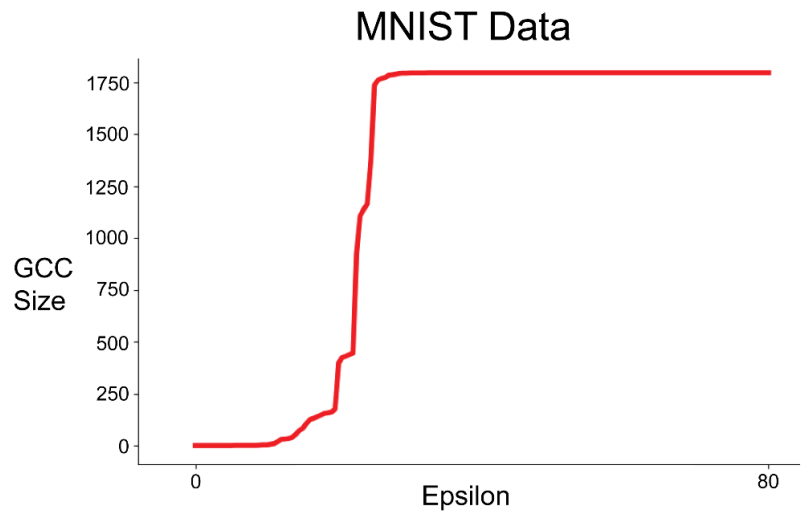

**Fig. S15.**
Epsilon plotted against the giant connected component size for the MNIST handwritten digits data.

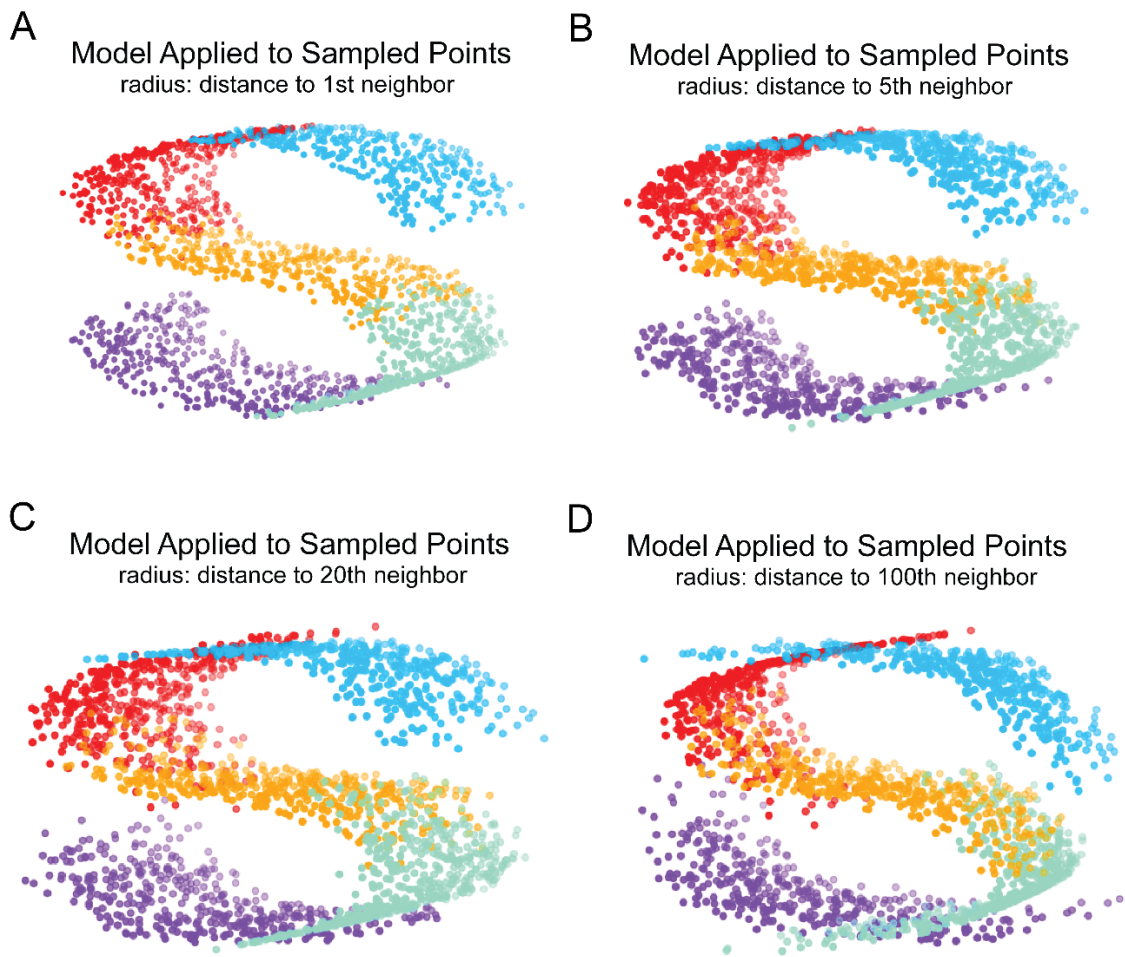

**Fig. S16.**
S-curves generated when applying the model to new data from sampled points. Data generated by sampling from the surface of a sphere with radius of the distance to the $1^{st}$, $5^{th}$, $20^{th}$, and $100^{th}$ nearest neighbor.